\documentclass[11pt]{article}

\usepackage{microtype}
\usepackage{graphicx}
\usepackage{subcaption}
\usepackage{booktabs}
\usepackage{microsoft-tech-report}
\usepackage{hyperref}
\usepackage{amsmath}
\usepackage{amssymb}
\usepackage{mathtools}
\usepackage{amsthm}
\usepackage{bm}
\usepackage{multirow}
\usepackage{enumitem}
\usepackage[capitalize,noabbrev]{cleveref}
\usepackage{xspace}
\usepackage{titletoc}
\usepackage{placeins}  % provides \FloatBarrier
\usepackage{tcolorbox} % colored boxes for theorems
\usepackage[ruled,linesnumbered,noend,noline]{algorithm2e} % pseudocode
\DontPrintSemicolon
\usepackage{listings}  % for code/JSON in prompt boxes
\usepackage{hyperref}
\usepackage{marvosym}
\tcbuselibrary{theorems,skins,breakable,listings}

\definecolor{promptbg}{HTML}{F7F7F8}
\definecolor{promptborder}{HTML}{5B5B5B}
\newtcolorbox{promptbox}[1][]{
  colback=promptbg, colframe=promptborder, fonttitle=\bfseries\sffamily,
  breakable, enhanced, arc=3pt, boxrule=0.8pt,
  left=10pt, right=10pt, top=8pt, bottom=8pt,
  title={#1}
}

\lstdefinestyle{json}{
  basicstyle=\small\ttfamily,
  breaklines=true,
  breakatwhitespace=false,
  showstringspaces=false,
  columns=fullflexible,
  literate={"}{\textquotedbl}1,
}

\newif\ifappendixtoc
\appendixtoctrue   % <-- change to \appendixtocfalse to disable

\definecolor{color_blue}{HTML}{E7EFFA}
\definecolor{color_green}{HTML}{E6F8E0}
\definecolor{color_gray}{HTML}{ECECEC}
\definecolor{pearDark}{HTML}{2980B9}
\definecolor{theoremblue}{HTML}{EBF5FB}
\definecolor{theoremborder}{HTML}{2980B9}
\definecolor{propgreen}{HTML}{EAFAF1}
\definecolor{propborder}{HTML}{27AE60}
\definecolor{defyellow}{HTML}{FEF9E7}
\definecolor{defborder}{HTML}{F39C12}
\definecolor{remarkgray}{HTML}{F2F3F4}
\definecolor{remarkborder}{HTML}{7F8C8D}

\hypersetup{
  colorlinks=true,
  linkcolor=msftblue,
  citecolor=msftblue,
  urlcolor=msftblue,
  pdftitle={[Your Paper Title]}          % <-- Replace with your title
}

\newtcbtheorem[auto counter]{theorem}{Theorem}{
  colback=theoremblue, colframe=theoremborder, fonttitle=\bfseries,
  breakable, enhanced, sharp corners, boxrule=0.6pt, left=6pt, right=6pt, top=6pt, bottom=6pt,
  crefname={Theorem}{Theorems}
}{thm}

\newtcbtheorem[use counter from=theorem]{proposition}{Proposition}{
  colback=propgreen, colframe=propborder, fonttitle=\bfseries,
  breakable, enhanced, sharp corners, boxrule=0.6pt, left=6pt, right=6pt, top=6pt, bottom=6pt,
  crefname={Proposition}{Propositions}
}{prop}

\newtcbtheorem[use counter from=theorem]{lemma}{Lemma}{
  colback=theoremblue, colframe=theoremborder, fonttitle=\bfseries,
  breakable, enhanced, sharp corners, boxrule=0.6pt, left=6pt, right=6pt, top=6pt, bottom=6pt,
  crefname={Lemma}{Lemmas}
}{lem}

\newtcbtheorem[use counter from=theorem]{corollary}{Corollary}{
  colback=propgreen, colframe=propborder, fonttitle=\bfseries,
  breakable, enhanced, sharp corners, boxrule=0.6pt, left=6pt, right=6pt, top=6pt, bottom=6pt,
  crefname={Corollary}{Corollaries}
}{cor}

\newtcbtheorem[use counter from=theorem]{definition}{Definition}{
  colback=defyellow, colframe=defborder, fonttitle=\bfseries,
  breakable, enhanced, sharp corners, boxrule=0.6pt, left=6pt, right=6pt, top=6pt, bottom=6pt,
  crefname={Definition}{Definitions}
}{def}

\newtcbtheorem[use counter from=theorem]{assumption}{Assumption}{
  colback=defyellow, colframe=defborder, fonttitle=\bfseries,
  breakable, enhanced, sharp corners, boxrule=0.6pt, left=6pt, right=6pt, top=6pt, bottom=6pt,
  crefname={Assumption}{Assumptions}
}{asm}

\newtcbtheorem[use counter from=theorem]{remark}{Remark}{
  colback=remarkgray, colframe=remarkborder, fonttitle=\bfseries,
  breakable, enhanced, sharp corners, boxrule=0.6pt, left=6pt, right=6pt, top=6pt, bottom=6pt,
  crefname={Remark}{Remarks}
}{rem}

\crefname{tcb@cnt@theorem}{Theorem}{Theorems}
\crefname{tcb@cnt@proposition}{Proposition}{Propositions}
\crefname{tcb@cnt@lemma}{Lemma}{Lemmas}
\crefname{tcb@cnt@corollary}{Corollary}{Corollaries}
\crefname{tcb@cnt@definition}{Definition}{Definitions}
\crefname{tcb@cnt@assumption}{Assumption}{Assumptions}
\crefname{tcb@cnt@remark}{Remark}{Remarks}
\crefname{algocf}{Algorithm}{Algorithms}

\techreportshorttitle{[Short Title]}      % <-- Replace with your short title

\begin{document}
\thispagestyle{empty}

% ============================================================
%  Header: Microsoft logo + date
% ============================================================
\noindent
\begin{minipage}[c]{0.5\linewidth}
\raggedright
\raisebox{-0.5\height}{\msftbrandmark}
\end{minipage}
\begin{minipage}[c]{0.49\linewidth}
\raggedleft
{\msftdatefont\small\color{msftgray}May, 2026}    % <-- Replace with your date
\end{minipage}\par
\vspace{0.35em}
\noindent{\color{msftline}\rule{\linewidth}{0.8pt}\par}

% ============================================================
%  Title + Authors
% ============================================================
\vspace{1.0em}
\begin{center}
{{\msfttitlefont\fontsize{21}{25}\selectfont\color{msftdark}
Lens: Rethinking Training Efficiency for \\Foundational Text-to-Image Models\par}}                         % <-- Replace with your title
\vspace{1.25em}

{\normalsize\rmfamily\color{msftdark}
\hyperref[sec:contributor]{Microsoft Lens Team}\par

%{\normalsize\rmfamily\color{msftdark}
%Microsoft Lens-ImageGen Team\par
% Usage:
%   $^{number}$   = affiliation
%   $^{*}$        = equal contribution
%   $^{\dagger}$  = internship or special note
%   $^{\ddagger}$ = corresponding author
% Add \\[-0.1em] between lines if you have many authors
}

\end{center}

% ============================================================
%  Abstract box
% ============================================================
\vspace{0.45em}
\begin{msfttitlebox}
\setlength{\parindent}{0cm}
\setlength{\parskip}{0cm}
%\raggedright
%\nohyphens

\begin{abstract}
Training foundational text-to-image (T2I) models typically requires massive computational resources. We introduce \textit{Lens}, a 3.8B-parameter T2I model that achieves performance competitive with, and in several cases surpassing, state-of-the-art models with more than 6B parameters across various benchmarks, while requiring significantly less training compute. For example, \textit{Lens} requires only about 19.3\% of the training compute used by Z-Image. The training efficiency of \textit{Lens} stems from two key strategies beyond its compact model size. First, we maximize \textit{data information density per training batch} by (i) training on \textit{Lens-800M}, a dataset of 800M densely captioned image-text pairs whose captions are generated by GPT-4.1 and contain approximately 109 words on average, providing richer semantic supervision than conventional short captions, and (ii) constructing each batch from images with multiple resolutions and diverse aspect ratios, thereby enlarging the effective visual coverage of each optimization step. Second, we improve \textit{convergence speed} through careful architectural choices, including adopting a semantic VAE that provides better latent representations and employing a strong language encoder that accelerates optimization while enabling multilingual generalization from English-only training data. After pre-training, we apply reinforcement learning with taxonomy-driven prompts (\textit{Lens-RL-8K}) and structured reward rubrics to suppress artifacts and improve visual quality, a reasoner module with training-free system prompt search to better align user requests with the model, and distillation-based acceleration for 4-step inference. Through efficient training and systematic optimization, \textit{Lens} generalizes to arbitrary aspect ratios from $1{:}2$ to $2{:}1$ and resolutions up to $1440^2$, and supports prompts in several commonly used languages. Thanks to its compact size, \textit{Lens} generates a $1024^2$ image in 3.15 seconds on a single NVIDIA H100 GPU, while its distilled turbo version performs 4-step generation in 0.84 seconds.
\end{abstract}

\vspace{0.14cm}
{\setlength{\parskip}{0.06cm}\small
% Uncomment and fill in the metadata fields you need:
{\msftmetalabel{Project Page}\href{https://github.com/microsoft/Lens}{https://github.com/microsoft/Lens}\par}
{\msftmetalabel{Date}May, 2026\par}                % <-- Replace with your date
}

\end{msfttitlebox}
\suppressfloats[t]  % Prevent floats from appearing above the abstract on page 1

% ============================================================
%  Optional teaser figure (uncomment and replace)
% ============================================================
% \begin{figure}[t]
%   \centering
%   \includegraphics[width=\textwidth]{figures/teaser.pdf}
%   \caption{Your teaser figure caption.}
%   \label{fig:teaser}
% \end{figure}

% ============================================================
%  Main content
% ============================================================

\begin{figure}[!t]
    \centering
    \includegraphics[width=0.99\textwidth]{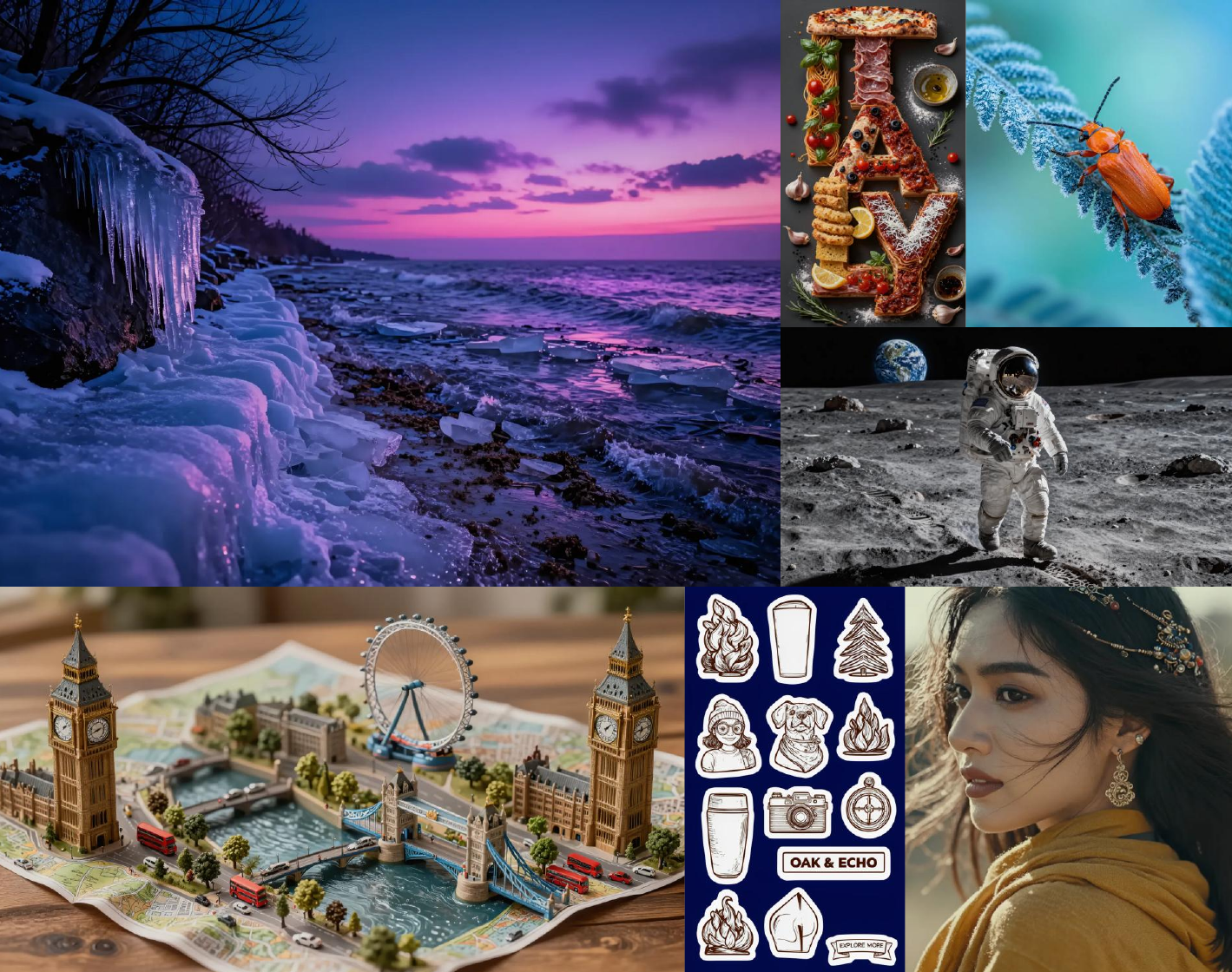}
    \caption{
    Images generated by \textit{Lens} at 1440 resolution. Section~\ref{sec:generation-visualization} provides more visualizations.
    }
    \label{fig:teaser}
\end{figure}

\section{Introduction}

Recent advances in foundational text-to-image (T2I) generative models have demonstrated remarkable capabilities in high-fidelity image synthesis and complex prompt understanding, as discussed in Appendix~\ref{sec:related works}. However, these gains have come at a substantial cost: training such models typically requires massive computational resources, leading to prohibitive financial and environmental expenses. For example, Z-Image~\citep{cai2025zimage} requires approximately 314K H800 GPU hours for pre-training, highlighting the growing scalability challenge of training foundation-scale T2I models.

In this paper, we focus on improving the \textbf{training-time efficiency} of foundational T2I models. We argue that training-time efficiency is jointly determined by three key factors: (1) \textit{model size}, which directly affects the computational cost of each training step; (2) \textit{data information density per training batch}, which determines how much useful supervision the model can extract from each update; and (3) \textit{convergence speed}, which determines the overall number of training iterations, as faster convergence enables the model to achieve strong performance with fewer optimization steps. Therefore, improving training-time efficiency requires not only reducing model scale, but also increasing the learning value of each batch and accelerating convergence throughout training.

\begin{figure*}[t]
  \centering
  \makebox[\textwidth][c]{%
  \begin{minipage}[t]{0.48\textwidth}
    \centering
    \includegraphics[width=\linewidth]{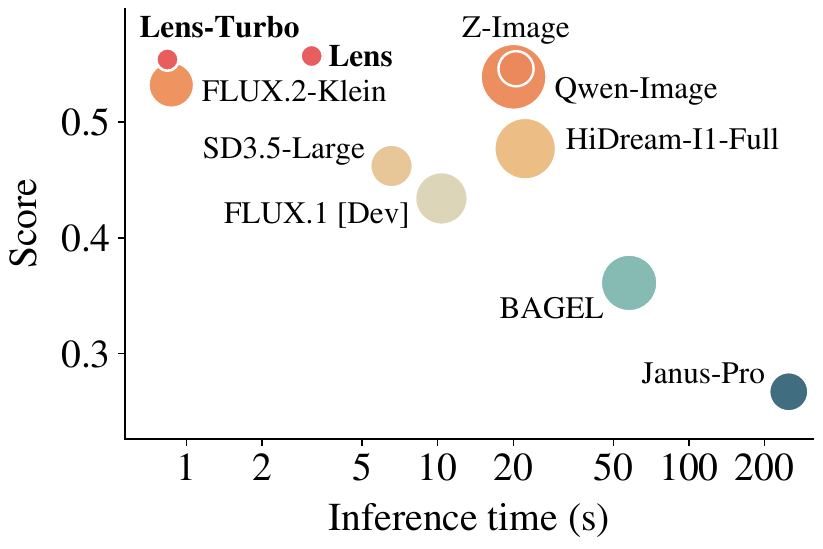}
    \\[-0.35em]
    \small (a) OneIG
  \end{minipage}
  \hspace{0.01\textwidth}
  \begin{minipage}[t]{0.48\textwidth}
    \centering
    \includegraphics[width=\linewidth]{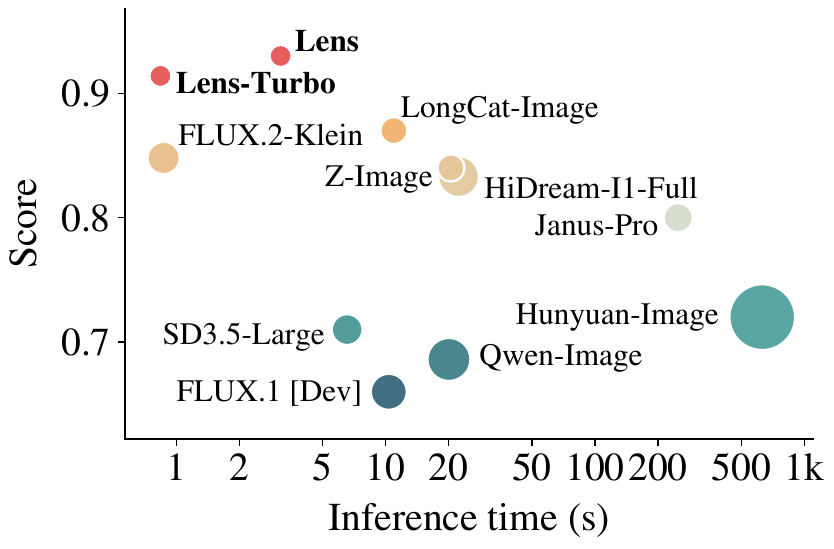}
    \\[-0.35em]
    \small (b) GenEval
  \end{minipage}%
  }
  %\vspace{-3mm}
  \caption{Comparison of inference time and benchmark performance on OneIG~\citep{chang2025oneigbench} and GenEval~\citep{ghosh2023geneval} across representative T2I models. The $x$-axis denotes inference time on a single NVIDIA H100 GPU, the $y$-axis denotes the benchmark score, and the marker area is proportional to model size.}
  %\vspace{-3mm}
  \label{fig:benchmark_scatter}
\end{figure*}

Motivated by these factors, we introduce \textbf{Lens}, a foundational T2I model designed for efficient training. First, to reduce the per-step computational cost, we constrain \textit{Lens} to 3.8B parameters. In contrast, recent state-of-the-art open-source models, including Z-Image (6B)~\citep{cai2025zimage}, LongCat-Image (6B)~\citep{team2025longcat}, FLUX.2 (9B)~\citep{flux-2-2025}, Qwen-Image (20B)~\citep{wu2025qwenimage}, and Hunyuan-Image-3.0 (MoE, 80B)~\citep{cao2025hunyuanimage}, operate at scales of 6B parameters or larger. Despite its relatively compact 3.8B-parameter scale, \textit{Lens} achieves performance competitive with, and in several cases surpassing, prior state-of-the-art larger models across multiple benchmarks, as shown in Figure~\ref{fig:benchmark_scatter}, while substantially reducing training cost. For example, compared with Z-Image (6B)~\citep{cai2025zimage}, \textit{Lens} (3.8B) attains competitive or superior results while using only approximately 19.3\% of its training compute.
%, computed as $(192\times312)/(314\times989.5)\approx0.193$. 
Specifically, \textit{Lens} requires 192K A100 GPU hours (312 TFLOPS, BF16), whereas Z-Image requires 314K H800 GPU hours (989.5 TFLOPS, BF16).\footnote{This comparison uses peak BF16 TFLOPS to normalize GPU types. Actual efficiency may differ due to memory bandwidth, MFU, and communication overhead. Re-captioning costs are excluded, as this one-time preprocessing can be reused for future models.} Moreover, due to its smaller model size, \textit{Lens} also enables faster inference under the same number of denoising steps.

Despite its reduced model size, the high training efficiency and strong performance of \textit{Lens} are largely attributed to two additional factors: \textit{data information density per training batch} and \textit{convergence speed}.

\noindent\textbf{Data Information Density per Training Batch.}
Given a training batch consisting of a set of \textit{image-text pairs}, our objective is to maximize the amount of useful visual-semantic supervision contained in each optimization step. To this end, we increase information density from both the text and image perspectives:

\begin{itemize}[leftmargin=*, labelsep=0.3em]
    \item \textbf{Text Information Density.} Conventional short captions provide limited supervision, as they often describe only the most salient object or scene category. In contrast, dense captions encode richer semantic details, including objects, attributes, spatial relationships, actions, and background context, allowing each image-text pair to provide stronger training signals. This effectively increases the text-side information density of the dataset. Accordingly, \textit{Lens} is trained on 800M densely captioned image-text pairs, where each caption is generated by a strong vision-language model, \textit{GPT-4.1}, with an average length of 109 words.
    \item \textbf{Image Information Density.} We increase image-side information density by constructing each training batch from images with multiple resolutions (i.e., $\{512^2, 768^2, 1024^2\}$) and diverse aspect ratios (e.g., $1{:}2$, $9{:}16$, $1{:}1$, and $4{:}3$). This strategy significantly increases image information density within each training batch: multi-resolution training allows the model to learn visual content at different levels of detail, from global scene structure to fine local patterns, while multi-aspect-ratio training exposes it to diverse object arrangements, spatial relationships, and compositional layouts. Moreover, a useful by-product of this strategy is strong \textit{resolution and aspect-ratio generalization} at inference time: the model generalizes well to unseen aspect ratios (e.g., $5{:}4$ and $6{:}7$) and to resolutions up to $1440^2$. This capability removes the need for costly high-resolution training, which further enhances overall training efficiency when high-resolution generation is desired. 
\end{itemize}

\noindent\textbf{Convergence Speed.} \textit{Lens} further enhances training efficiency by accelerating optimization convergence. We explore several architectural design choices that allow the model to learn more effectively and reach strong performance with fewer training iterations. These studies include:

\begin{itemize}[leftmargin=*, labelsep=0.3em]
\item \textbf{VAE Variants.} We systematically study different VAE variants, including conventional VAEs used in FLUX.1~\citep{flux2024} and SD3~\citep{esser2024scaling}, as well as semantic VAEs adopted in FLUX.2~\citep{flux-2-2025} and VTP~\citep{vtp}. Instead of relying on proxy metrics such as rFID or class-conditional ImageNet generation, we directly evaluate each VAE within the T2I pipeline using a 130M subset of our training data. 

\item \textbf{Language Encoder Variants.} The language encoder provides text-conditioning features for diffusion modeling. We find that stronger language encoders not only accelerate optimization convergence but also improve multilingual generalization. Specifically, although the model is trained only on English image-text pairs, a strong language encoder enables robust inference-time generalization to other languages, such as Chinese and French. This multilingual generalization substantially reduces data requirements and training costs in scenarios where the model needs to handle multilingual inputs. Based on careful ablation studies, we adopt GPT-OSS~\citep{openai2025gptoss120bgptoss20bmodel} as the language encoder.
\end{itemize}

After efficient pre-training, \textit{Lens} generates diverse images, but their aesthetic quality may vary and some outputs may contain artifacts. We apply reinforcement learning (RL) as a post-training step to suppress artifacts, improve visual composition, and enforce consistency with real-world physical rules. A key finding is that RL data must be sufficiently diverse and cover the original training distribution to avoid performance degradation on certain input types. To this end, we construct the \textit{Lens-RL-8K} prompt set with taxonomy-driven coverage of diverse generation scenarios. Experiments show that post-training on \textit{Lens-RL-8K} significantly improves generation performance across a broad range of scenarios.

Additionally, following modern T2I systems, we equip \textit{Lens} with a \textit{reasoner} module that can be instantiated with different LLMs. The reasoner converts ambiguous or underspecified user requests into detailed prompts aligned with the training-caption distribution. It takes the user request and a system prompt as input, where the system prompt specifies guidelines for constructing suitable T2I prompts. We further introduce a \textit{training-free system prompt search strategy} to optimize these guidelines, enabling the reasoner to generate prompts that better align with the T2I model. Note that reasoner-based prompt rewriting is now a standard practice in modern T2I systems; to ensure a fair comparison, we report results both with and without the reasoner in our experiments.

Overall, in this paper we systematically investigate a set of training efficiency factors that are often overlooked in practice, including data captioning strategies, VAE selection criteria, language encoder choices, and training-data composition for RL-based post-training. For each factor, we provide controlled ablation studies with quantitative analysis, yielding actionable insights for building T2I foundation models. Importantly, these strategies are complementary to conventional training acceleration approaches, such as architectural innovations and distributed-system optimization. Guided by these findings, \textit{Lens} achieves performance competitive with larger state-of-the-art models at substantially lower training cost. Its compact model size also enables faster inference: by default, \textit{Lens} generates a $1024^2$ image in \textbf{3.15} seconds on a single NVIDIA H100 GPU using 20 denoising steps, while \textit{Lens-Turbo}, a 4-step distilled variant, further reduces the generation time to \textbf{0.84} seconds.

\section{Method}

\begin{figure*}[t]
  \centering
  \makebox[\textwidth][c]{%
  \begin{minipage}[t]{0.29\textwidth}
    \centering
    \includegraphics[width=\linewidth]{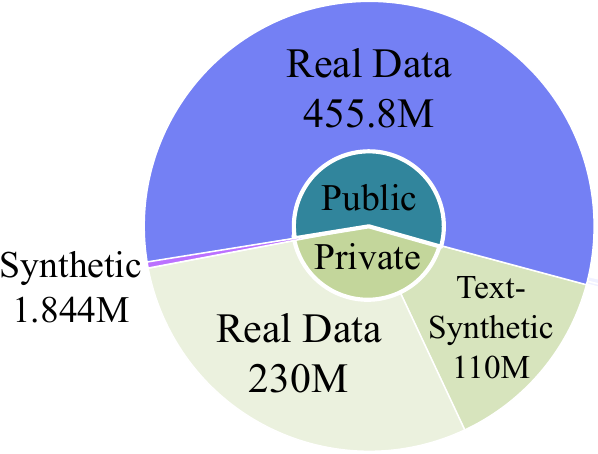}
    \\[-0.05em]
    \small (a) Lens-800M.
  \end{minipage}
  \hspace{0.001\textwidth}
  \begin{minipage}[t]{0.3\textwidth}
    \centering
    \includegraphics[width=\linewidth]{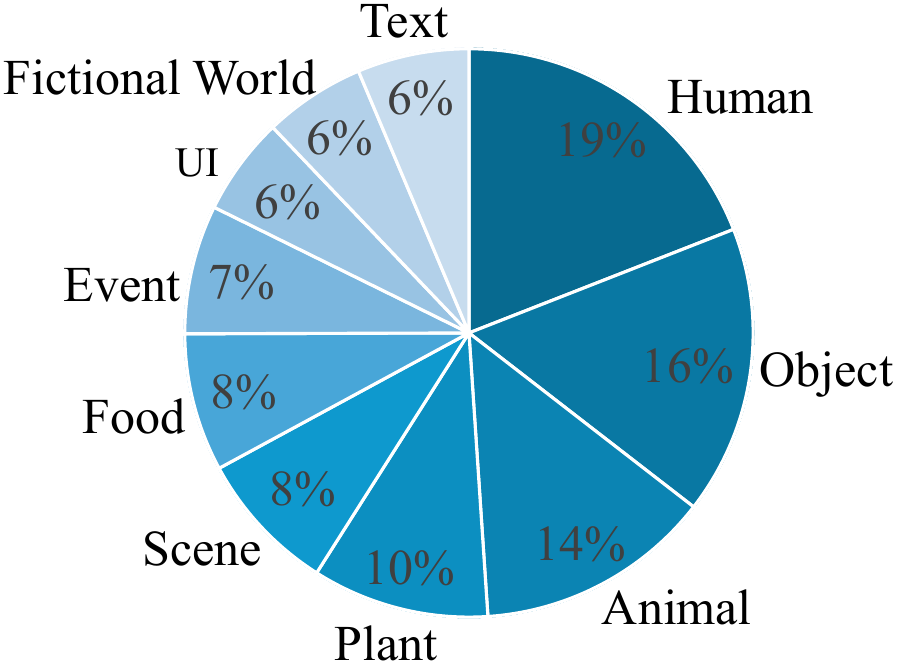}
    \\[-0.05em]
    \small (b) Lens-RL-8K.
  \end{minipage}%
  \hspace{0.001\textwidth}
  \begin{minipage}[t]{0.38\textwidth}
    \centering
    \includegraphics[width=\linewidth]{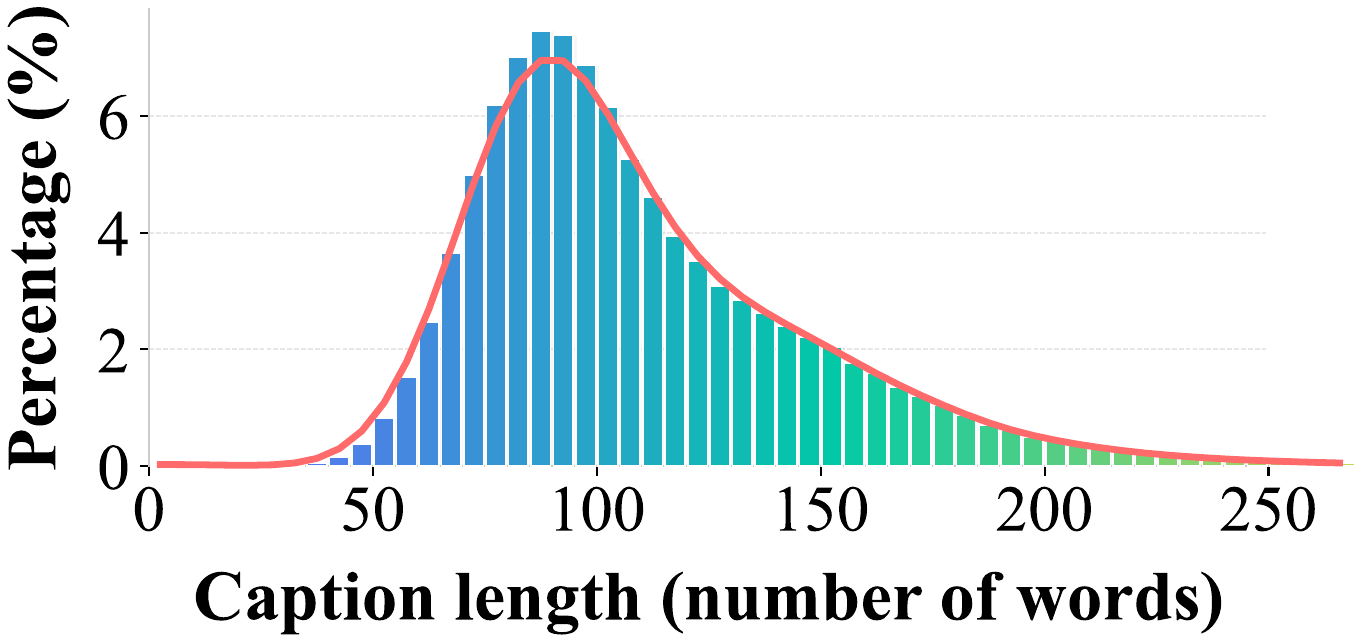}
    \\[-0.05em]
    \small (c) Lens-800M.
  \end{minipage}%
  }
  %\vspace{-2mm}
  \caption{Distribution of (a) the pre-training dataset, \textit{Lens-800M}, (b) the RL dataset, \textit{Lens-RL-8K}, and (c) caption length distribution of the \textit{Lens-800M} dataset, caption length is measured by the number of words, with an average length of approximately 109 words.}
  %\vspace{-3mm}
  \label{fig:data-distribution}
\end{figure*}

In this section, we present the details of \textit{Lens}. We first describe the construction of the training dataset, \textit{Lens-800M}, in Section~\ref{sec:data}. We then present the model architecture in Section~\ref{sec:archi}, followed by the pre-training recipe in Section~\ref{sec:pre-training}. In Section~\ref{sec:post-training}, we introduce our RL-driven post-training strategy, which is built on the carefully designed \textit{Lens-RL-8K} dataset and optimized reward rubrics. We further introduce few-step distillation to distill \textit{Lens} into \textit{Lens-Turbo}, a 4-step generator that does not require CFG. Finally, Section~\ref{sec:optimization} discusses inference configuration and training-free system-prompt search.

\begin{figure}[t]
    \centering
    \begin{minipage}[t]{0.55\linewidth}
    \includegraphics[width=1.0\linewidth]{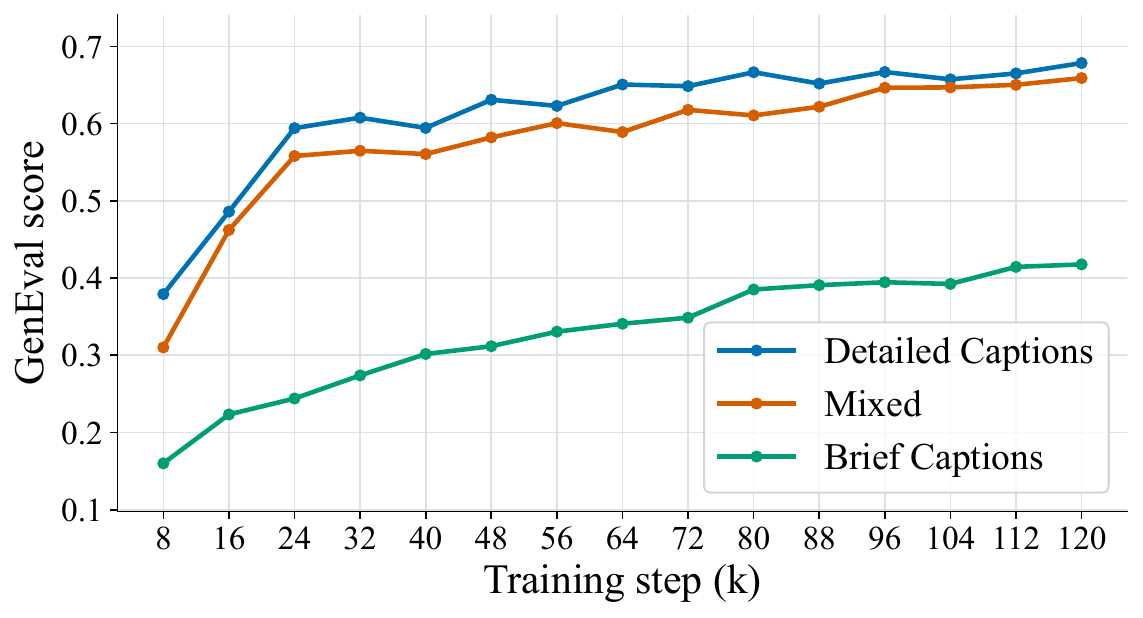}
    \vspace{-18pt}
    \caption{Caption-length ablation study.}
    \label{fig:abl_caption}
    \end{minipage}
\hfill
    \begin{minipage}[t]{0.44\linewidth}
    \includegraphics[width=1.0\linewidth]{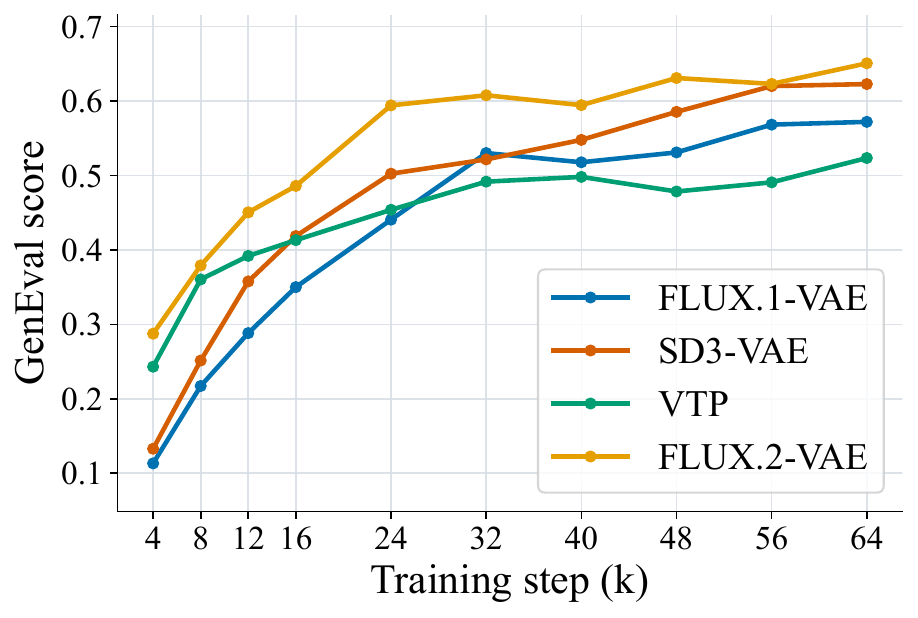}
    \vspace{-18pt}
    \caption{Ablation study on VAE variants.}
    \label{fig:abl_vae}
    \end{minipage}
    \vspace{-10pt}
\end{figure}

\subsection{Pre-training Data: Lens-800M}
\label{sec:data}

\noindent\textbf{Data Distribution.}
Our pre-training corpus is constructed from four complementary sources to ensure content diversity: \textit{(1) public real-world data}; \textit{(2) public synthetic data}; \textit{(3) private data}, covering text-heavy visual content such as posters, slides, graphic designs, and general-domain images; and \textit{(4) text synthetic data}, where text is rendered onto randomly sampled backgrounds with augmentations in blur, color, font, scale, and rotation to increase typographic and layout diversity. 

We apply a multi-stage data-cleaning pipeline to ensure the quality of the \textit{Lens-800M} pre-training dataset:
(1) removing corrupted or broken files;
(2) resolution filtering, where images with an area smaller than $384^2$ are removed;
(3) NSFW content filtering using an EVA model~\citep{freepik2025nsfw} fine-tuned for NSFW classification;
(4) aesthetic filtering using Aesthetic Predictor v2.5~\citep{schuhmann2022laion}, where samples with scores below 3 are discarded;
(5) watermark filtering using a SigLIP2 model~\citep{tschannen2025siglip} fine-tuned for watermark detection;
(6) clarity filtering, where visually blurry samples are removed based on the variance of the Laplacian computed on scale-normalized grayscale images;
(7) entropy filtering, where low-information samples are removed based on the Shannon entropy of grayscale intensity histograms;
(8) luminance filtering, where under- or over-exposed samples are removed based on the mean V-channel value in HSV color space, normalized to $[0,1]$;
and (9) near-duplicate removal using CLIP ViT-L/14 embeddings with a cosine-similarity threshold of $>0.985$, accelerated by FAISS~\citep{douze2024faiss, johnson2019billion} indexing.

After the data filtering process, the final pre-training dataset contains approximately \textit{800M} high-quality images. The detailed data distribution is illustrated in Figure~\ref{fig:data-distribution}(a). We refer to this pre-training dataset as \textbf{Lens-800M}.

\noindent\textbf{Captioning Images with Detailed Captions.}
For each image in \textit{Lens-800M}, we employ a strong vision-language model, GPT-4.1 in our implementation, to generate a detailed, long-form \textit{English} caption using the prompt described in Appendix~\ref{sec:prompt-caption}. At the same time, to preserve multilingual rendering capabilities, any text appearing in the image is kept in its original language in the caption. Figure~\ref{fig:data-distribution}(c) presents the caption length statistics. Training examples are provided in Appendix~\ref{sec:lens-800M visualization}.

This design is motivated by three considerations.
(1) \textit{Improving data quality.} Web-crawled alt-text captions are often short, underspecified, and sometimes incorrect. Such noisy supervision forces the model to resolve ambiguity during training, leading to inefficient capacity usage and degraded learning signals~\citep{betker2023improving,chen2024sharegpt4v}.
(2) \textit{Bridging the training–inference gap.} In real-world usage, users frequently provide long and compositional prompts to describe desired images. Training on detailed captions better aligns the model with this inference-time distribution.
(3) \textit{Enhancing data efficiency.} Empirically, we observe that training \emph{exclusively} on dense captions yields the best generation performance, outperforming short-caption training.

\noindent\textbf{Ablation Study: Detailed vs. Brief Captions.}
To validate observation (3), we conduct a controlled ablation study. We randomly sample 130M images from the \textit{Lens-800M} dataset to construct an ablation subset, denoted as \textbf{Lens-130M}. We train three small text-to-image models (referred to as \textbf{Lens-Toy}) with identical architectures (described in Section~\ref{sec:archi}), each using a 1.2B-parameter image generation backbone and a Qwen3-0.6B text encoder. The only difference lies in the captioning strategy:
\textit{(i) Brief}: where GPT-4.1 generates short and sparse captions (e.g., ``a photo of a cat'') for each image in \textit{Lens-130M};
\textit{(ii) Detailed}, which uses our generated dense captions; and \textit{(iii) Mixed}, a 50/50 combination of Brief and Detailed captions. We evaluate generation performance on the GenEval~\citep{ghosh2023geneval} benchmark. As shown in Figure~\ref{fig:abl_caption}, training with dense captions achieves better generation quality than the other variants, owing to improved data utilization efficiency.

\begin{tcolorbox}[colback=gray!5,colframe=black]
\textit{Dense caption supervision improves the data information density of each training batch, leading to better data utilization and higher training efficiency.}
\end{tcolorbox}

\subsection{Architecture}
\label{sec:archi}

Our model mainly consists of: (1) a \textit{VAE} that encodes images into compact latents; (2) a \textit{Latent Diffusion Transformer} that denoises text-conditioned image latents; (3) a \textit{Reasoner} that converts ambiguous user requests into detailed, well-formed prompts.

\noindent\textbf{VAE.} We examine both classical VAEs, including those used in FLUX.1~\citep{flux2024} and SD3~\citep{esser2024scaling}, and semantic VAEs, including those used in FLUX.2~\citep{flux-2-2025} and VTP~\citep{vtp}. We do not use rFID to evaluate VAE performance, since reconstruction fidelity mainly measures how well a VAE reproduces a given image, rather than how effectively its latent space supports generative learning. We also avoid relying on class-conditional ImageNet generation as a proxy evaluation. Instead, we directly assess each VAE in the text-to-image generation setting by training \textit{Lens-Toy} models on the \textit{Lens-130M} dataset introduced in Section~\ref{sec:data}. As shown in Figure~\ref{fig:abl_vae}, FLUX.2's VAE achieves the best generation performance while also accelerating model convergence, and is therefore adopted as the VAE in \textit{Lens}.

\begin{tcolorbox}[colback=gray!5,colframe=black]
\textit{The VAE is crucial for both generation quality and model convergence. A strong VAE defines a more compact and semantically meaningful visual latent space, making text-image alignment easier to learn and reducing the number of optimization steps required for convergence.}
\end{tcolorbox}

\begin{figure}[t]
  \centering
  \includegraphics[width=\textwidth]{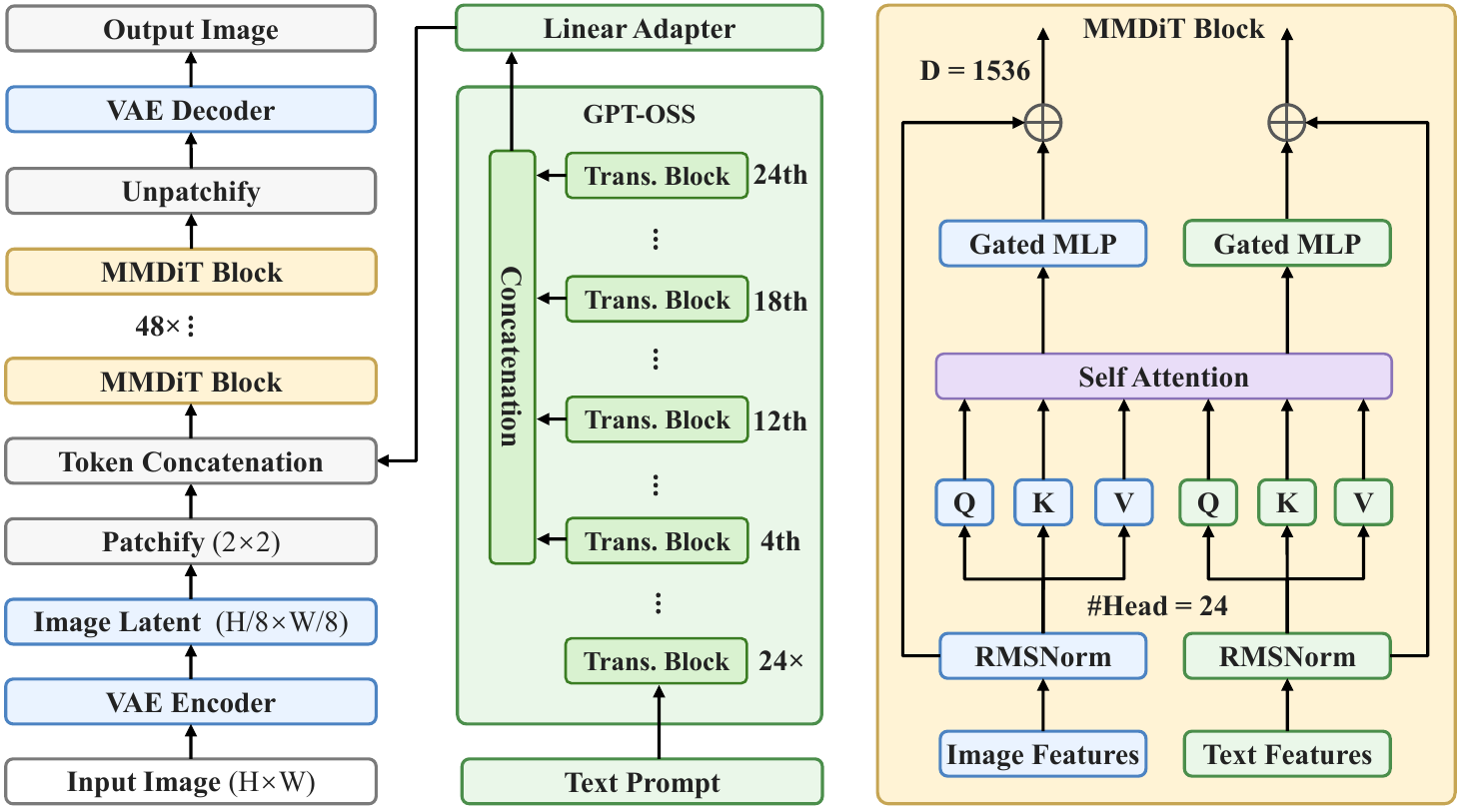}
 % \vspace{-6mm}
  \caption{Architecture of the latent diffusion Transformer in \textit{Lens} (left) and the detailed design of an MMDiT block (right). ``Trans.'': Transformer. }
  %\vspace{-3mm}
  \label{fig:architecture}
\end{figure}

\noindent\textbf{Latent Diffusion Transformer.}
We adopt an MMDiT-style~\citep{esser2024scaling} architecture for \textit{Lens}, as illustrated in Figure~\ref{fig:architecture}. \textit{Lens} is formulated as a latent diffusion model and trained with the standard flow-matching~\citep{lipman2023flow} objective. Image latents are extracted using the FLUX.2 VAE, while text features are obtained from GPT-OSS~\citep{openai2025gptoss120bgptoss20bmodel}, a 20B-parameter MoE language model with 3B activated parameters and 24 layers in total. To better leverage multi-level semantic representations, we extract GPT-OSS features from the 4th, 12th, 18th, and 24th layers and concatenate them along the feature dimension. A linear adapter is then applied to project the concatenated text representation into the same dimensionality as the image latents. The denoising backbone consists of 48 MMDiT blocks. Each block takes as input the concatenation of noisy image features and the text features produced by the previous MMDiT block, and processes them through two separate branches for image and text modalities. We use RMSNorm~\citep{zhang2019root} as the normalization layer and apply RoPE~\citep{su2024roformer} to the image features.

\begin{figure}[t]
    \centering
    \begin{minipage}[t]{0.49\linewidth}
    \includegraphics[width=1.0\linewidth]{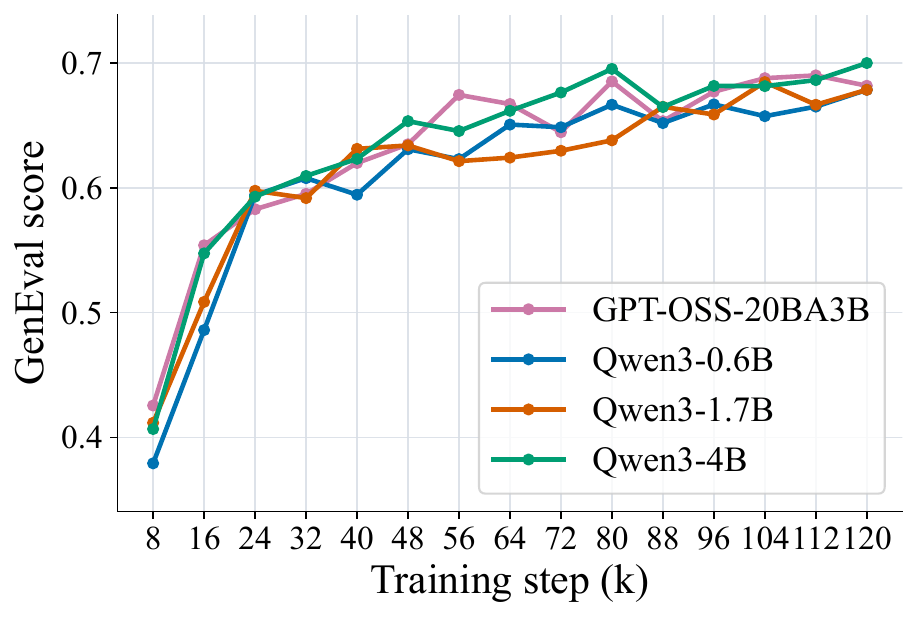}
    \vspace{-17pt}
    \caption{Study of different language encoders for English text-conditioned generation.}
    \label{fig:abl_lang_eng}
    \end{minipage}
\hfill
    \begin{minipage}[t]{0.49\linewidth}
    \includegraphics[width=1.0\linewidth]{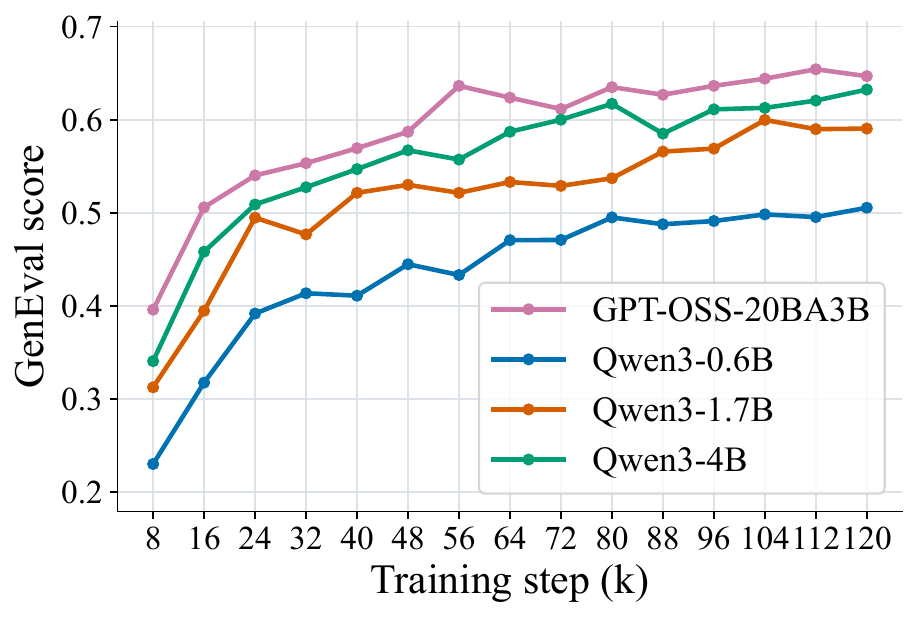}
    \vspace{-17pt}
    \caption{Study of language encoders for multilingual text-conditioned generation, averaged over five common languages (EN/ZH/FR/JA/ES).}
    \label{fig:abl_lang_multi}
    \end{minipage}
    %\vspace{-3mm}
\end{figure}

\noindent\textbf{Ablation Study: Language Encoder.}
We consider two key factors when selecting the language encoder: (1) whether a stronger language encoder can facilitate text-image alignment, leading to better generation performance and faster convergence; and (2) whether it can enable multilingual generalization, i.e., training on English-only image-text pairs while supporting inference in other languages. To verify these effects, we compare four language encoders: GPT-OSS (MoE, 20B-A3B)~\citep{openai2025gptoss120bgptoss20bmodel} and Qwen3~\citep{yang2025qwen3} with different model sizes, including 0.6B, 1.7B, and 4B. We use \textit{Lens-Toy} as the ablation model and construct four variants that differ only in the choice of language encoder. All variants are trained on \textit{Lens-130M}, which contains 130M image-text pairs with English-only captions. Figures~\ref{fig:abl_lang_eng} and~\ref{fig:abl_lang_multi} present performance curves on the GenEval~\citep{ghosh2023geneval} benchmark as a function of training iterations. Based on these results, we adopt GPT-OSS as our language encoder.

\begin{tcolorbox}[colback=gray!5,colframe=black]
\textit{Strong language encoders provide a richer semantic text space for text-image alignment, bringing three key benefits: improved prompt-following fidelity, faster model convergence, and, most importantly, support for multilingual inputs beyond the training language without additional multilingual image-text training data.}
\end{tcolorbox}

\noindent\textbf{Reasoner.} The Reasoner is an independent language module placed before the T2I model. Its role is to interpret the user's raw input, refine ambiguous or underspecified instructions, and convert them into more detailed, coherent prompts optimized for generation. Because it functions independently of the T2I model's internal text encoder, the Reasoner can be easily swapped out without retraining the generation backbone. While we use GPT-5.5 as our default, the Reasoner is compatible with various commercial and open-source LLMs. Our evaluations in Appendix~\ref{sec:various reasoners} show that even using an open-source model like GPT-OSS provides substantial gains. This setup is particularly efficient: since GPT-OSS already functions as our text encoder, employing it as the Reasoner adds zero extra GPU memory cost, demonstrating that our framework can achieve superior results without relying on costly commercial APIs.

\subsection{Pre-training}
\label{sec:pre-training}

\noindent\textbf{Low-resolution Pre-training.} We first pre-train \textit{Lens} at a fixed resolution of $512\times512$ on 128 NVIDIA A100 80GB GPUs for 400K iterations. The FLUX.2 VAE and GPT-OSS language encoder are kept frozen, and only the diffusion transformer is optimized using the flow-matching MSE objective. We adopt logit-normal timestep sampling with $\mu{=}1.06$, corresponding to the 1024 image tokens of a $512\times512$ image. Training is performed in bfloat16 with gradient checkpointing. We use AdamW~\citep{loshchilov2019adamw} with $\beta_1{=}0.9$ and $\beta_2{=}0.999$, a constant learning rate of $2\times10^{-4}$, an effective global batch size of 3072 images, and gradient clipping set to 1.0.

\noindent\textbf{Mixed-resolution Continual Training.} Starting from the low-resolution checkpoint, we continue training for another 400K iterations using WebDataset bucket sampling over mixed resolutions on 128 NVIDIA A100 80GB GPUs. Specifically, we construct the resolution bucket set from three base image areas, $512^2$, $768^2$, and $1024^2$, combined with nine aspect ratios: $1{:}2$, $9{:}16$, $2{:}3$, $3{:}4$, $1{:}1$, $4{:}3$, $3{:}2$, $16{:}9$, and $2{:}1$. This results in 27 concrete resolution buckets: for the $512^2$ base, $352\times704$, $384\times672$, $416\times640$, $448\times608$, $512\times512$, $608\times448$, $640\times416$, $672\times384$, and $704\times352$; for the $768^2$ base, $544\times1088$, $576\times1024$, $640\times960$, $672\times896$, $768\times768$, $896\times672$, $960\times640$, $1024\times576$, and $1088\times544$; and for the $1024^2$ base, $736\times1472$, $768\times1376$, $832\times1248$, $864\times1152$, $1024\times1024$, $1152\times864$, $1248\times832$, $1376\times768$, and $1472\times736$.

For logit-normal timestep sampling, we adapt $\mu$ according to the image token length $n$: $\mu(n)$ is linearly interpolated from $\mu{=}1.0$ at $n{=}256$ tokens to $\mu{=}1.3$ at $n{=}4096$ tokens. We keep the same frozen VAE/language-encoder setup and optimizer as in low-resolution pre-training, and use per-base bucket batch sizes of 24, 10, and 6 for $512^2$, $768^2$, and $1024^2$, respectively. Since different ranks may process different resolutions within the same optimization step and high-resolution buckets require more computation, these resolution-dependent batch sizes are chosen to balance per-step wall-clock time across ranks. We train with a constant learning rate of $1\times10^{-4}$, while keeping the remaining optimizer configuration unchanged.

\noindent\textbf{Resolution and Aspect-ratio Generalization after Pre-training.}
Although the base model is trained on only 27 resolutions, constructed from 3 base areas and 9 aspect ratios, it generalizes well to unseen resolutions and aspect ratios at inference time. Specifically, it can generate images with arbitrary aspect ratios ranging from $1{:}2$ to $2{:}1$ and image areas up to $1440^2$, even though training does not include resolutions between $1024^2$ and $1440^2$, nor aspect ratios outside the predefined bucket set. This suggests that mixed-resolution pre-training does not simply encourage the model to memorize a fixed set of resolution buckets. Instead, exposure to diverse spatial scales and aspect ratios enables the model to learn more continuous and resolution-aware image representations. Moreover, the use of RoPE-based positional encoding may further facilitate such generalization, as it represents positions in a relative and extrapolatable manner.

\begin{tcolorbox}[colback=gray!5,colframe=black]
\textit{Mixed-resolution training enables strong resolution generalization at inference time, allowing the model to generate images at unseen sizes and diverse aspect ratios. This reduces the need to collect and train on images covering every target resolution and aspect-ratio combination. More importantly, since training on higher-resolution images increases computational cost quadratically with image size, resolution generalization allows the model to produce higher-resolution outputs while being trained primarily on lower-resolution images. This substantially reduces training cost and improves overall training efficiency.}
\end{tcolorbox}

\subsection{Post-training}
\label{sec:post-training}

After pre-training, our base model, \textit{Lens-Base}, can strictly follow user prompts and generate diverse images. However, the generated images may still contain visual artifacts. To further improve generation quality and reliability, we adopt reinforcement learning as a post-training strategy.

\noindent\textbf{Lens-RL-8K Dataset.} \textit{Lens-RL-8K} is a prompt dataset designed for RL-based post-training, consisting of $8{,}406$ prompts that cover a broad range of T2I generation scenarios. A key observation in this work is that RL prompts should match the generation-scenario distribution of the pre-training data as comprehensively as possible. This enables post-training to improve the model’s overall generation quality and alignment across diverse scenarios, rather than overfitting to a narrow set of prompt types.

We propose a taxonomy-driven construction pipeline for building the \textit{Lens-RL-8K} prompt set for RL training. First, we summarize common generation scenarios into a category set, including \texttt{Human}, \texttt{Object}, \texttt{Animal}, \texttt{Plant}, \texttt{Scene}, \texttt{Food}, \texttt{Event}, \texttt{Fictional World}, \texttt{Text}, and \texttt{UI and Graphic Design}. For each category, we further define dozens of fine-grained sub-categories. For example, the \texttt{Human} category includes sub-categories such as \texttt{Race}, \texttt{Occupation}, and \texttt{Gender}. Each sub-category then contains hundreds of concrete items. For instance, the \texttt{Race} sub-category includes items such as \texttt{White People} and \texttt{Asian People}, while the \texttt{Occupation} sub-category includes items such as \texttt{Researcher} and \texttt{Doctor}. In total, we construct an item set containing $8,406$ concrete items.

Next, we define a description set that specifies the key dimensions used to enrich each prompt, including \texttt{Attribute}, \texttt{Spatial Relationship}, \texttt{Count}, \texttt{Interaction}, and \texttt{Color}. These description dimensions provide basic guidance for generating diverse and detailed prompts for each concrete item.

Finally, for each item in the item set, we randomly sample one to four dimensions from the description set and prompt GPT-4.1 to generate an image-generation prompt using the system prompt detailed in Appendix~\ref{sec:prompt-lens-RL-8K}. This process yields $8,406$ RL prompts, forming the \textit{Lens-RL-8K} dataset. The category distribution of the prompts is shown in Figure~\ref{fig:data-distribution}(b).

\noindent\textbf{Rubric Generation.}
Rubrics define the key aspects used to assess images generated by \textit{Lens}. Given a prompt $\mathcal{P}$ from the \textit{Lens-RL-8K} dataset, we first generate 10 sample-aware rubrics using GPT-4.1. Specifically, we feed $\mathcal{P}$ together with the system prompt described in Appendix~\ref{sec:prompt-rubric} into GPT-4.1 to produce prompt-specific rubrics. We further append a global rubric, i.e., ``\textit{Verify that the entire image is structurally coherent and physically plausible}''. As a result, each prompt is associated with a set of evaluation rubrics. We also provide several prompt-rubric examples in Appendix~\ref{vis:rubric}.

\begin{table*}[t]
    \centering
    \vspace{-5mm}
    \caption{Left: Comparison of different \textit{Lens-RL} variants trained on different subsets of \textit{Lens-RL-8K} on GenEval. Right: Comparison between \textit{Lens-RL} models trained on the full \textit{Lens-RL-8K} set and a subset excluding text prompts on two text-rendering benchmarks, CVTG and OneIG (EN).}
    \vspace{-2mm}
    \label{tab:rl_diversity_ablation}

    \begin{tabular}{c@{\hspace{1mm}}c}
        {
        \renewcommand{\arraystretch}{1.06}
        \begin{tabular}[t]{lc}
            \toprule
            RL Training Set & GenEval \\
            \midrule
            1/4 Full set & 0.916 \\
            1/2 Full set & 0.920 \\
            Full set  & 0.930 \\
            \bottomrule
        \end{tabular}
        }
        &
        {
        \renewcommand{\arraystretch}{0.94}
        \begin{tabular}[t]{lcccc}
            \toprule
            \multirow{2}{*}{\raisebox{-0.6ex}{RL Training Set}} & \multicolumn{3}{c}{CVTG} & OneIG (EN) \\
            \cmidrule(r){2-4} \cmidrule(l){5-5}
             & Avg. & NED & CLIP & Text \\
            \midrule
            Full set \textit{w/o} text & 0.832 & 0.928 & 0.795 & 0.946 \\
            Full set  & 0.869 & 0.951 & 0.814 & 0.960 \\
            \bottomrule
        \end{tabular}
        }
    \end{tabular}
\end{table*}

\noindent\textbf{DiffusionNFT with VLM as Reward Function.}
We adopt DiffusionNFT~\citep{zheng2025diffusionnft} to optimize \textit{Lens-Base}, using GPT-4.1-mini as the reward function, inspired by RubricRL~\citep{feng2025rubricrl}. Specifically, at each optimization step, we randomly sample 48 prompt-rubric pairs and generate 24 images at different resolutions for each prompt using the current policy model. We then feed each generated image, together with its corresponding rubrics, into GPT-4.1-mini. Guided by the system prompt described in Appendix~\ref{sec:prompt-reward}, GPT-4.1-mini produces rewards that are used by DiffusionNFT to optimize the policy model. We train the RL policy for 180 steps on 64 NVIDIA A100 GPUs. Further details are provided in Appendix~\ref{sec:RL-Details}.

\noindent\textbf{Ablation Study: Various RL Datasets.} To verify that prompt diversity in the RL dataset is crucial for post-training performance, we conduct two ablation studies. First, we compare our default model, \textit{Lens-RL}, trained on the full \textit{Lens-RL-8K} dataset, with a variant trained on \textit{Lens-RL-8K} after removing text-related prompts. Second, we compare \textit{Lens-RL} with variants trained on smaller subsets of \textit{Lens-RL-8K}, including one-half and one-quarter of the full dataset, which substantially reduce the diversity of RL training prompts. All models are initialized from the same base model and trained for the same 180 RL steps. The results are reported in Table~\ref{tab:rl_diversity_ablation}.

\begin{tcolorbox}[colback=gray!5,colframe=black]
\textit{A broad and well-balanced RL prompt set enables the model to improve across diverse generation scenarios, whereas reduced or biased prompt coverage limits generalization.}
\end{tcolorbox}

\noindent\textbf{Few-step Distillation.} To improve sampling efficiency, we distill \textit{Lens-RL} into \textit{Lens-Turbo}, a 4-step generator distilled from a curated, well-balanced image-caption dataset. Our recipe combines effective techniques from DMD2~\citep{dmd2}, decoupled-DMD~\citep{decoupleddmd}, and SenseFlow~\citep{senseflow}, together with R1 regularization~\citep{R1penalty, apt} on the adversarial loss to improve training stability. \textit{Lens-Turbo} largely preserves the image quality and prompt-following ability of the original model. Details are provided in Appendix~\ref{sec:Distill-Details}.

\subsection{Inference}
\label{sec:optimization}

By default, we first use a reasoner to refine ambiguous or underspecified user inputs into more detailed prompts. The refined prompts are then fed into \textit{Lens} for \textbf{20}-step generation with CFG set to 5.0. For faster inference, \textit{Lens-Turbo} performs \textbf{4}-step generation without CFG.

\noindent\textbf{Training-free System-prompt Search.}
This technique aims to find a better system prompt for the reasoner, enabling it to more effectively convert user requests into suitable T2I model prompts. It requires no model training. Instead, we iteratively feed the previous system prompt and generation analysis, i.e., textual summaries of failure cases, into GPT-5.5, and ask it to rewrite the system prompt. We find that this training-free strategy produces improved system prompts for the reasoner. This strategy is not specific to our T2I system; it can also be applied to other T2I models.

\section{Comparison with State-of-the-art Models}

\begin{table*}[!t]
  \centering
  \caption{Comparison of \textit{Lens} with 20-step inference and \textit{Lens-Turbo} with 4-step inference against state-of-the-art models across four benchmarks. We report overall scores for \textit{OneIG}, \textit{GenEval}, and \textit{LongText (EN)}, and average score, normalized edit distance (NED), and CLIP score for \textit{CVTG}. Detailed benchmark comparisons are provided in Tables~\ref{tab:main_oneig},~\ref{tab:main_geneval}, and~\ref{tab:main_longtext_cvtg} in Appendix~\ref{sec:benchmark-details}. The best open-source results are highlighted in \textbf{bold}, and the second-best results are \underline{underlined}.}
  \vspace{-1mm}
  \label{tab:main_merged}
  \begin{tabular}{l|c|cc|c|c|ccc}
    \toprule
    \multicolumn{1}{l|}{\multirow{2}{*}{\raisebox{-0.6ex}{Model}}}
      & \multicolumn{1}{c|}{\multirow{2}{*}{\raisebox{-0.6ex}{Size}}}
      & \multicolumn{2}{c|}{OneIG}
      & \multicolumn{1}{c|}{\multirow{2}{*}{\raisebox{-0.6ex}{GenEval}}}
      & \multicolumn{1}{c|}{\multirow{2}{*}{\raisebox{-3ex}{\shortstack[c]{LongText\\(EN)}}}}
      & \multicolumn{3}{c}{CVTG} \\
    \cmidrule(lr){3-4}\cmidrule(l){7-9}
      & & EN & ZH & & & Avg. & NED & CLIP \\
    \midrule
    \multicolumn{9}{l}{\emph{Commercial models}} \\
    Kolors 2.0 & --
      & 0.434 & 0.426 & -- & 0.258 & -- & -- & -- \\
    Seedream 3.0 & --
      & 0.530 & 0.528 & 0.843 & 0.896 & 0.592 & 0.854 & 0.782 \\
    Seedream 4.0 & --
      & 0.573 & 0.554 & 0.840 & 0.921 & 0.892 & 0.951 & 0.785 \\
    GPT Image 1 [High] & --
      & 0.533 & 0.474 & 0.840 & 0.956 & 0.857 & 0.948 & 0.798 \\
    Nano Banana 2.0 & --
      & 0.578 & 0.567 & -- & 0.981 & -- & -- & -- \\
    \midrule
    \multicolumn{9}{l}{\emph{Open-source models}} \\
    Janus-Pro & 7B
      & 0.267 & 0.240 & 0.800 & 0.019 & -- & -- & -- \\
    BAGEL & 14B
      & 0.361 & 0.370 & -- & 0.373 & -- & -- & -- \\
    HiDream-I1-Full & 17B
      & 0.477 & 0.337 & 0.833 & 0.543 & -- & -- & -- \\
    SD3.5 Large & 8B
      & 0.462 & -- & 0.710 & -- & 0.655 & 0.847 & 0.780 \\
    FLUX.1 [Dev] & 12B
      & 0.434 & -- & 0.660 & 0.607 & 0.496 & 0.688 & 0.740 \\
    FLUX.2-Klein & 9B
      & 0.532 & 0.430 & 0.848 & 0.864 & -- & -- & -- \\
    Z-Image-Turbo & 6B
      & 0.528 & 0.507 & 0.823 & 0.917 & 0.859 & 0.928 & 0.805 \\
    Z-Image & 6B
      & 0.546 & \underline{0.535} & 0.840 & 0.935 & 0.867 & 0.937 & 0.797 \\
    Qwen-Image & 20B
      & 0.539 & \textbf{0.548} & 0.868 & \textbf{0.943} & 0.829 & 0.930 & 0.806 \\
    Hunyuan-Image-3.0 & 80B
      & -- & -- & 0.720 & -- & 0.765 & 0.877 & 0.812 \\
    LongCat-Image & 6B
      & -- & -- & 0.870 & -- & 0.866 & 0.936 & 0.786 \\
    \midrule
    \textbf{Lens-Turbo} (4-step) & \textbf{3.8B}
      & \underline{0.554} & 0.519 & \underline{0.914} & 0.927 & \textbf{0.889} & \textbf{0.965} & \textbf{0.815} \\
    \textbf{Lens} (20-step) & \textbf{3.8B}
      & \textbf{0.557} & 0.525 & \textbf{0.930} & \underline{0.937} & \underline{0.869} & \underline{0.951} & \underline{0.814} \\
    \bottomrule
  \end{tabular}
\end{table*}

\noindent\textbf{Main Results}. In Table~\ref{tab:main_merged}, we evaluate \textit{Lens} against state-of-the-art models on four text-to-image benchmarks:

\noindent\textbf{(1) OneIG (EN)}~\citep{chang2025oneigbench} is the English split of OneIG-Bench, comprising 1,120 prompts across general objects, portraits, anime/stylization, text rendering, and knowledge/reasoning. It evaluates generated images with dimension-specific scores for subject alignment, text accuracy, reasoning, style, diversity, and overall performance. 

\noindent\textbf{(2) GenEval}~\citep{ghosh2023geneval} focuses on object-centric compositional alignment, with 553 prompts spanning six templated tasks: single-object generation, two-object co-occurrence, counting, color, spatial relation, and attribute binding. It uses detector- and classifier-based verification to assess whether generated images satisfy these structured constraints.

\noindent\textbf{(3) LongText (EN)}~\citep{geng2025xomni} contains 160 English prompts across eight text-rich scenarios, including signboards, labeled objects, printed materials, webpages, slides, posters, captions, and dialogues. Its prompts include both short text of roughly 10--30 words and longer text of 30--50 words, stressing faithful rendering beyond isolated words or phrases. 

\noindent\textbf{(4) CVTG}~\citep{du2025textcrafter} evaluates complex visual text generation with multiple text regions, where prompts vary the number of regions from 2 to 5 and specify text content, position, length, and style attributes such as color, font, and size.

\section{Visualization}
\label{sec:generation-visualization}

\begin{figure}[!t]
    \centering
    \includegraphics[width=0.95\textwidth]{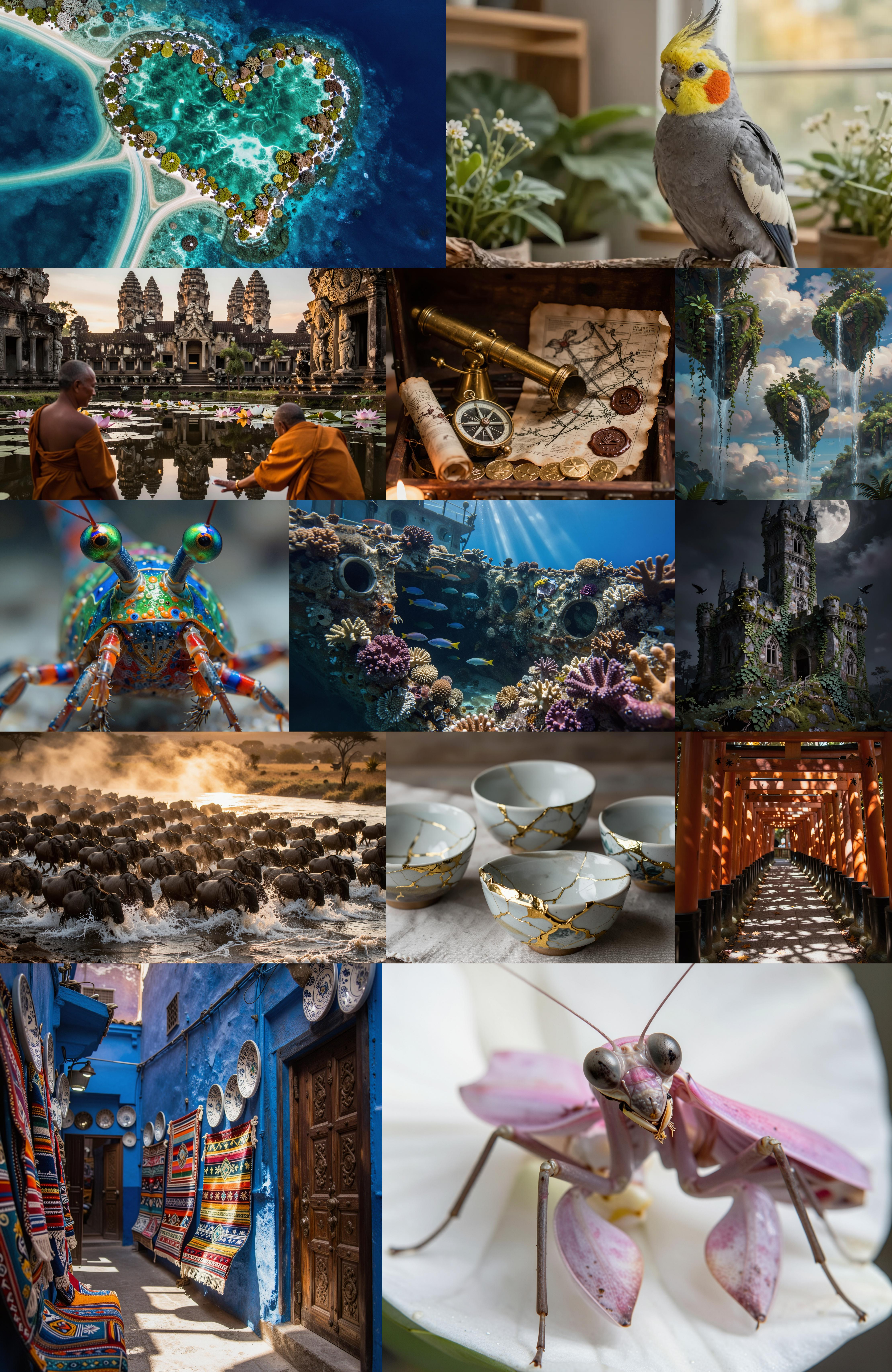}
    \caption{
  General image gallery. \textit{Lens} generates diverse, high-resolution images across natural scenes, animals, architecture, objects, and imaginative worlds.
}
    \label{fig:general_01}
\end{figure}

\begin{figure}[!t]
    \centering
    \includegraphics[width=0.95\textwidth]{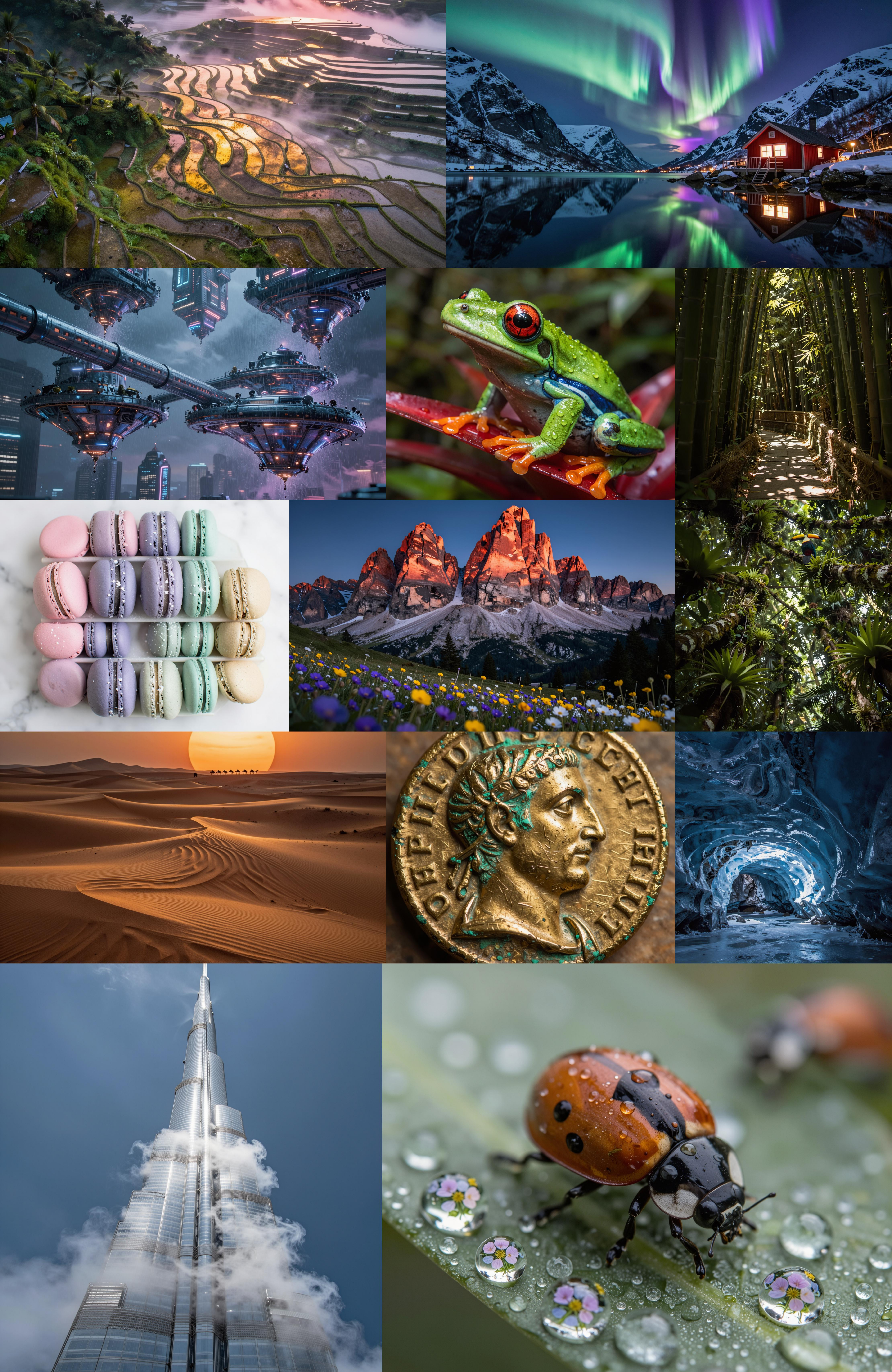}
    \caption{
  General image gallery, demonstrating broad visual diversity, fine-grained details, and strong aesthetic quality across multiple domains.
}
    \label{fig:general_02}
\end{figure}

\begin{figure}[!t]
    \centering
    \includegraphics[width=0.95\textwidth]{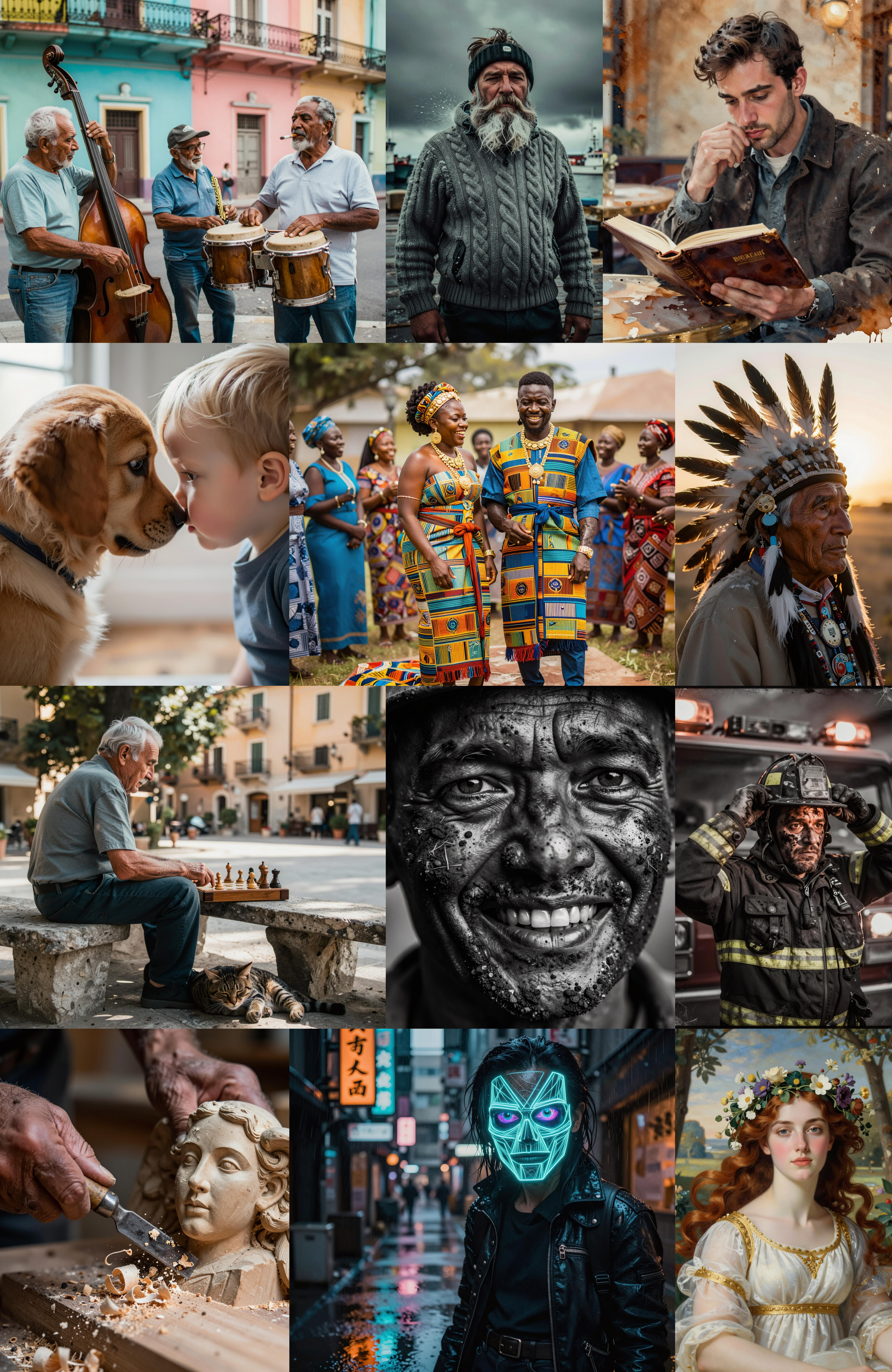}
    \caption{
  Portrait gallery. \textit{Lens} captures diverse human subjects with expressive details, natural lighting, and rich contextual storytelling.
}
    \label{fig:portrait_01}
\end{figure}

\begin{figure}[!t]
    \centering
    \includegraphics[width=0.95\textwidth]{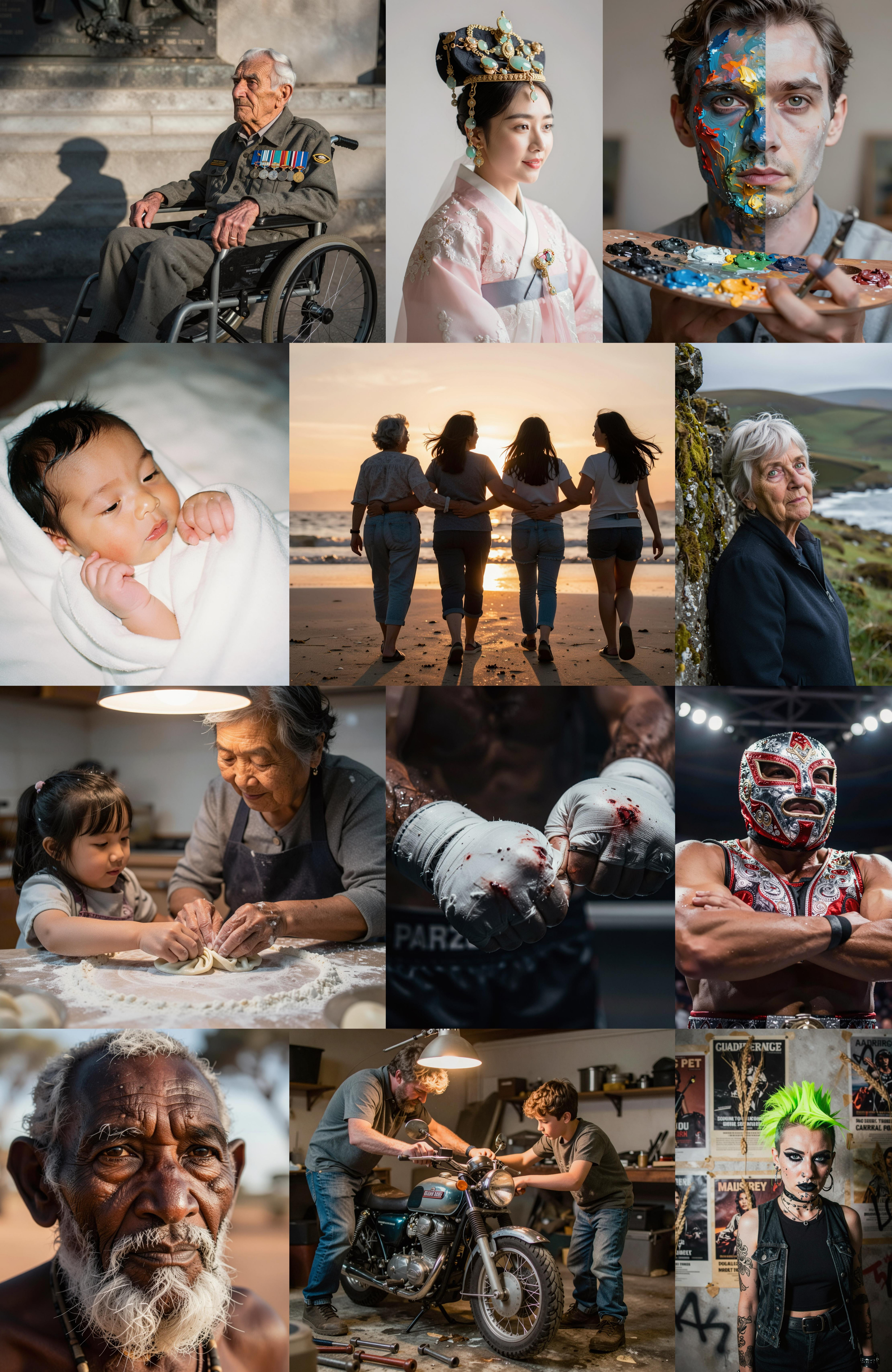}
    \caption{
  Generated portrait samples showcasing identity diversity, fine-grained facial details, cinematic composition, and varied cultural and narrative settings.
}
    \label{fig:portrait_02}
\end{figure}

\begin{figure}[!t]
    \centering
    \includegraphics[width=0.95\textwidth]{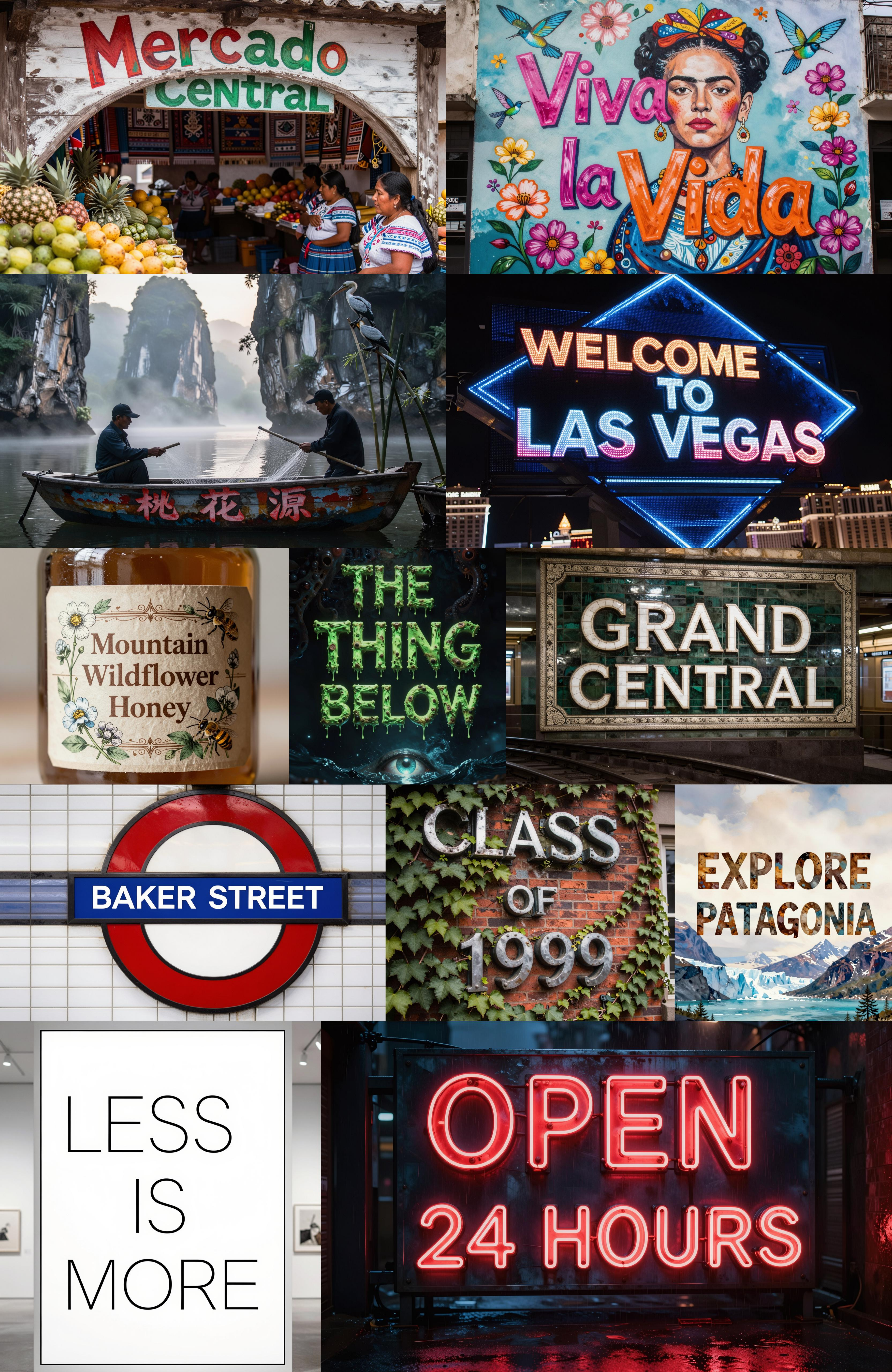}
    \caption{
  Text rendering samples, demonstrating the model’s ability to generate legible typography across posters, signs, product labels, and stylized graphic designs.
}
    \label{fig:text_01}
\end{figure}

\begin{figure}[!t]
    \centering
    \includegraphics[width=0.95\textwidth]{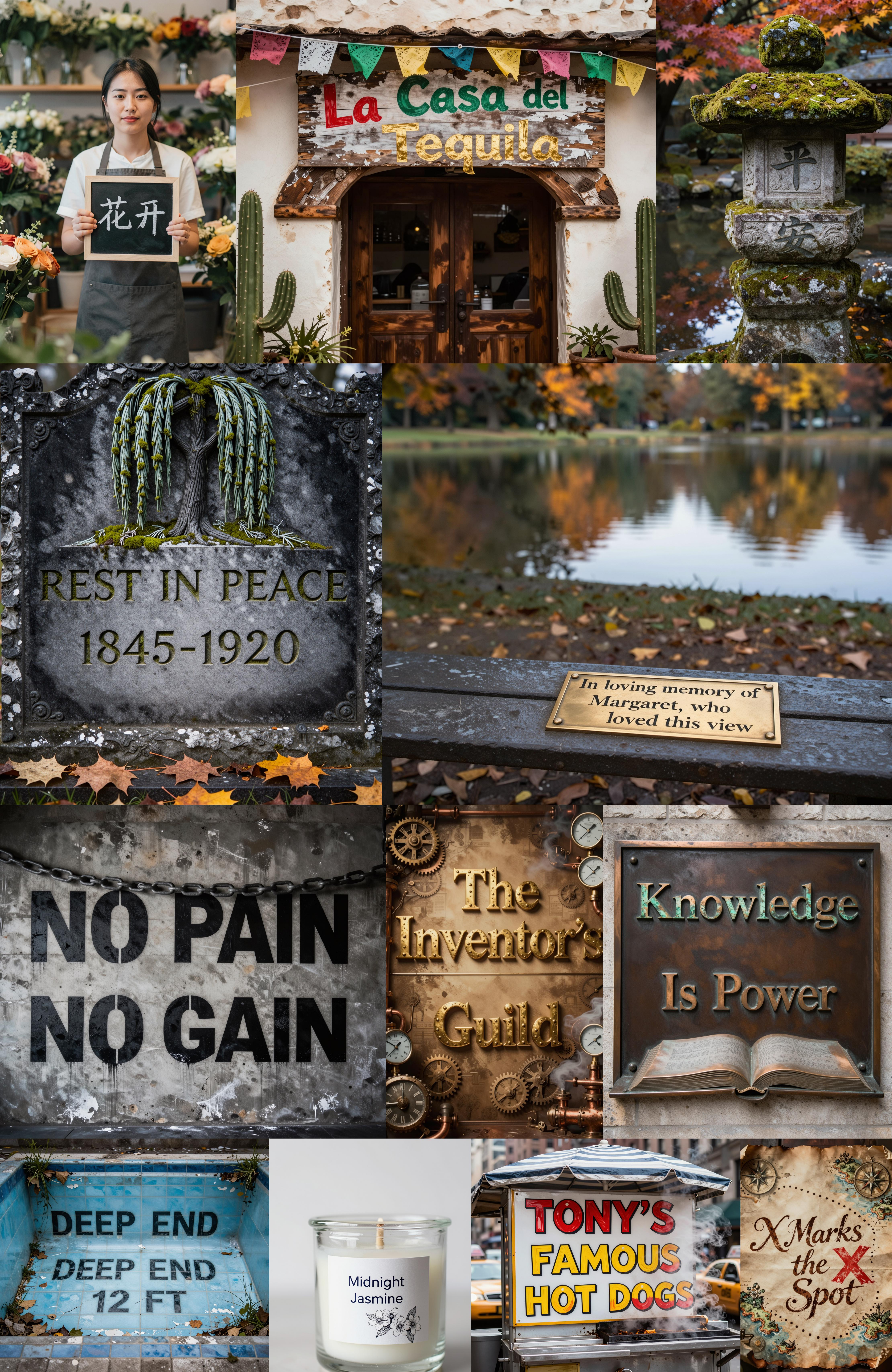}
    \caption{
  Additional text-rich generations, covering diverse visual contexts from storefronts and posters to labels, murals, and environmental signage.
}
    \label{fig:text_02}
\end{figure}

\begin{figure}[!t]
    \centering
    \includegraphics[width=0.95\textwidth]{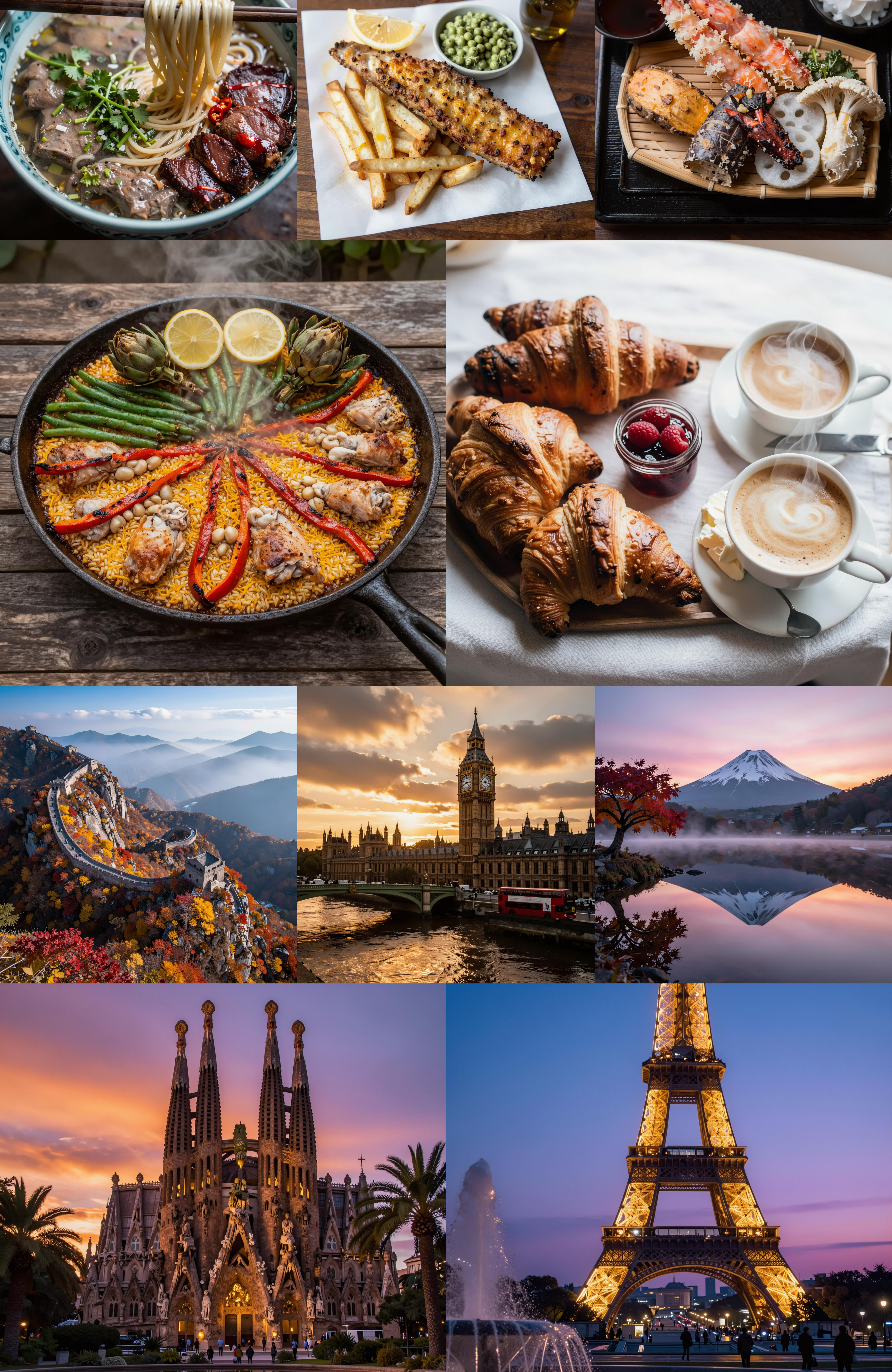}
    \caption{
  Multilingual prompt following on culturally representative cuisines and landmarks, covering regional foods, iconic architecture, and world-famous scenic destinations.
}
    \label{fig:multilingual_01}
\end{figure}

\begin{figure}[!t]
    \centering
    \includegraphics[width=0.75\textwidth]{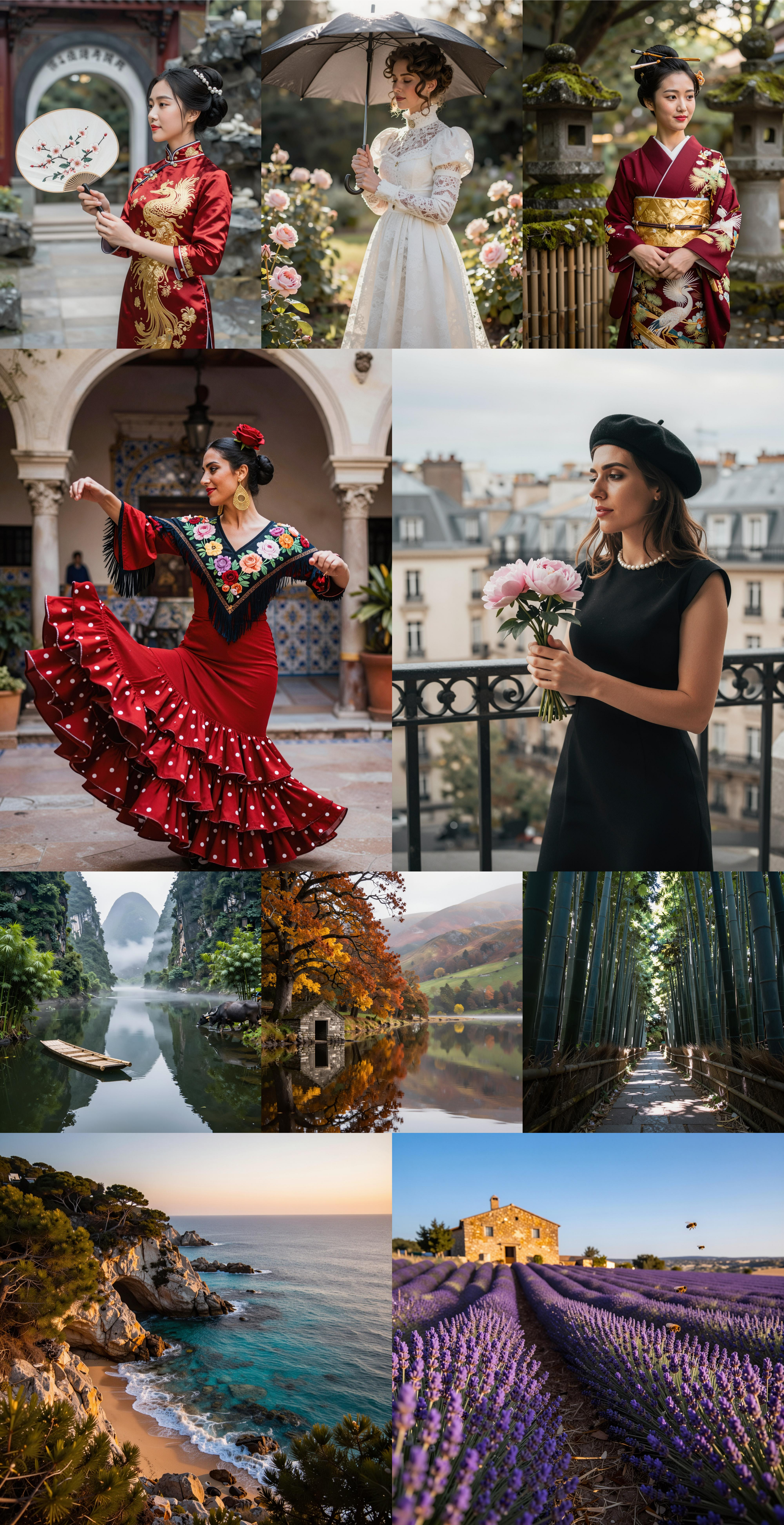}
    \caption{
  Multilingual prompt following on cultural identity and regional scenery, including traditional clothing, natural landscapes, and language-specific visual contexts.
}
    \label{fig:multilingual_02}
\end{figure}

We present qualitative visualizations of images generated by \textit{Lens-Turbo} across diverse generation scenarios. Specifically, we show results for general image generation (Figures~\ref{fig:general_01} and~\ref{fig:general_02}), portrait generation (Figures~\ref{fig:portrait_01} and~\ref{fig:portrait_02}), multilingual visual text generation (Figures~\ref{fig:text_01} and~\ref{fig:text_02}), and multilingual prompt following (Figures~\ref{fig:multilingual_01} and~\ref{fig:multilingual_02}), where the input prompts are written in different languages. All generated images have an area of $1440^2$ pixels, with varying aspect ratios. These examples demonstrate the ability of \textit{Lens-Turbo} to produce high-quality images, render text in multiple languages, generate realistic portraits, and generalize to multilingual user instructions.

\section{Conclusion}

In this work, we introduce \textbf{Lens}, a 3.8B-parameter foundational T2I model designed for training-time efficiency. By improving data information density through dense captions and mixed-resolution/aspect-ratio pre-training on the proposed \textit{Lens-800M} dataset, and by accelerating convergence with carefully selected VAE and language encoder designs, \textit{Lens} achieves strong generation quality at substantially lower training cost. We further enhance the model through RL-based post-training on the diverse \textit{Lens-RL-8K} prompt set, together with system-level optimizations including a reasoner, training-free system-prompt search, and few-step distillation. Extensive experiments show that \textit{Lens} achieves performance competitive with, and in several cases surpassing, larger state-of-the-art models while enabling fast inference, demonstrating that efficient training strategies can substantially improve the scalability of foundational T2I models.

\section*{Contributor List (Alphabetical Order)}
\label{sec:contributor}

\noindent\textbf{Project Leads:} Dong Chen (doch@microsoft.com), Fangyun Wei (fawe@microsoft.com), Ziyu Wan (ziyuwan@microsoft.com)

\noindent\textbf{Core Contributors:} Dongdong Chen, Jiawei Zhang, Jinjing Zhao, Sirui Zhang, Yang Yue, Zhiyang Liang

\noindent\textbf{Contributors:} Baining Guo, Chong Luo, Jianmin Bao, Ji Li, Lei Shi, Qinhong Yang, Xuelu Feng, Xiuyu Wu, Yan Lu, Yanchen Dong, Yitong Wang, Yunuo Chen

\newpage
\FloatBarrier  % Ensure all floats appear before References
\bibliography{references}
\bibliographystyle{unsrtnat}

\newpage
\newpage
\appendix

This appendix includes: 
\begin{itemize}
    \item Related work discussion (Appendix~\ref{sec:related works}).
    \item Detailed comparisons of \textit{Lens} against state-of-the-art models on OneIG, GenEval, LongText (EN), and CVTG (Appendix~\ref{sec:benchmark-details}).
    \item Reasoner analysis (Appendix~\ref{sec:various reasoners}).
    \item Rubric visualizations and training-data visualizations (Appendix~\ref{sec:visualization-appendix}).
    \item Implementation details (Appendix~\ref{sec:implementation details}).
    \item All prompts used in this work (Appendix~\ref{sec:all-prompts}).
    \item Broader impacts and limitations of this work (Appendix~\ref{sec:limitations}).
\end{itemize}

\section{Related Works}
\label{sec:related works}

\subsection{Foundational T2I Models}
Text-to-image (T2I) foundation models have progressed from latent diffusion models to large-scale diffusion Transformers, rectified-flow models, and proprietary multimodal image-generation systems. Early latent diffusion models (LDMs), including Stable Diffusion, established the paradigm of performing generation in a compressed latent space, substantially reducing training and inference costs compared with pixel-space diffusion~\citep{rombach2022high}. Building on this paradigm, SDXL further improves high-resolution synthesis through a larger UNet, stronger text conditioning, multi-aspect-ratio training, and a dedicated refinement stage~\citep{podell2024sdxl}.

More recent models have shifted toward Transformer-based backbones and rectified-flow objectives. Stable Diffusion 3 and its 3.5 variants adopt MMDiT-style architectures with rectified-flow training~\citep{esser2024scaling,stabilityai2024sd35}, following the broader trend of scalable diffusion Transformers~\citep{peebles2023scalable}. Recent open-source and commercial systems, such as FLUX~\citep{flux2024}, SANA~\citep{xie2024sana}, HiDream-I1~\citep{cai2025hidream}, Qwen-Image~\citep{wu2025qwenimage}, Hunyuan-Image-3.0~\citep{cao2025hunyuanimage}, Z-Image~\citep{cai2025zimage}, GPT Image~\citep{openai2025gptimage}, Gemini 2.5 Flash Image (Nano Banana)~\citep{google2025nanobanana}, Kolors 2.0~\citep{kolors2}, and Seedream 4.0~\citep{seedream2025seedream4}, further advance visual quality, prompt following, text rendering, image editing, reference consistency, and inference efficiency.

In parallel, unified multimodal and autoregressive generators aim to integrate image synthesis with language modeling and visual understanding. Janus-Pro~\citep{chen2025januspro} scales a unified autoregressive framework for both multimodal understanding and generation. Transfusion~\citep{zhou2024transfusion} combines next-token prediction with diffusion over continuous image representations, while BAGEL~\citep{deng2025BAGEL} scales decoder-only multimodal pretraining on interleaved text, image, video, and web data. Although these approaches achieve strong performance and broaden the capabilities of T2I systems, they often require increasingly large models, massive training data, and complex system designs. This trend motivates our study of efficient foundation-model training, aiming to achieve competitive generation quality with substantially reduced computational cost.

\subsection{Post-training for T2I Models}

Post-training has become an important stage for improving the alignment and generation quality of T2I models. Early efforts mainly relied on supervised fine-tuning, while recent work has increasingly explored preference-based optimization and reinforcement learning (RL). These methods aim to improve model behavior beyond likelihood-based pre-training by directly optimizing for human preferences, reward-model judgments, or task-specific generation criteria.

One major line of work adapts direct preference optimization (DPO) to diffusion models. These methods train the model using positive-negative image pairs or preference sets, encouraging generations preferred by humans or reward models while suppressing less desirable outputs. Representative approaches include Diffusion-DPO~\citep{wallace2024diffusion}, D3PO~\citep{yang2024using}, SPO~\citep{liang2025aesthetic}, and related variants~\citep{yuan2024self,yang2024dense,li2024aligning,hong2026margin}. Compared with online RL, DPO-style methods are relatively simple and stable, as they optimize preference objectives without requiring explicit reward maximization during sampling.

Another line of work applies policy-gradient-based RL to the generation process. GRPO-based methods formulate image generation as a sequential decision-making problem and optimize the denoising or flow-matching trajectory directly. For flow-matching models, Flow-GRPO~\citep{liu2025flow} and its variants~\citep{li2025mixgrpo,wang2025pref} extend policy-gradient optimization to continuous generative dynamics by converting deterministic sampling into an equivalent stochastic formulation. Related methods, such as DiffusionNFT~\citep{zheng2025diffusionnft} and AWM~\citep{xue2025advantage}, further investigate RL objectives over the generative process, introducing online optimization frameworks and advantage-aware or negative-aware objectives to guide model improvement.

Across both preference-optimization and RL-based post-training, reward design is a critical factor. A useful reward should capture not only global aesthetic quality, but also prompt faithfulness, object correctness, spatial and compositional consistency, text rendering, and safety-related constraints. Poorly designed rewards may cause reward hacking, reduced diversity, or misalignment between the optimization objective and human preference. Therefore, recent work~\citep{feng2025rubricrl,gunjal2025rubrics,huang2025reinforcement,he2025advancedif} has increasingly emphasized fine-grained, multi-dimensional reward construction. This motivates our rubric-based post-training strategy, which provides explicit and structured evaluation criteria for RL optimization of T2I foundation models.

\begin{table*}[t]
  \centering
  \caption{Comparison of \textit{Lens} with commercial and open-source models on \textbf{OneIG (EN)}. The best results are highlighted in \textbf{bold}, and the second-best results are \underline{underlined}.}
  \vspace{1mm}
  \label{tab:main_oneig}
  \small
  \setlength{\tabcolsep}{4.8pt}
  \renewcommand{\arraystretch}{1.12}
  \resizebox{\textwidth}{!}{%
  \begin{tabular}{l|c|cccccc}
    \toprule
    Model & Size & Alignment & Text & Reasoning & Style & Diversity & Overall \\
    \midrule
    \multicolumn{8}{l}{\emph{Commercial models}} \\
    Kolors 2.0 & --
      & 0.820 & 0.427 & 0.262 & 0.360 & 0.300 & 0.434 \\
    Seedream 3.0 & --
      & 0.818 & 0.865 & 0.275 & 0.413 & 0.277 & 0.530 \\
    Seedream 4.0 & --
      & 0.892 & 0.983 & 0.347 & 0.453 & 0.191 & 0.573 \\
    GPT Image 1 [High] & --
      & 0.851 & 0.857 & 0.345 & 0.462 & 0.151 & 0.533 \\
    Nano Banana 2.0 & --
      & 0.888 & 0.944 & 0.334 & 0.481 & 0.245 & 0.578 \\
    \midrule
    \multicolumn{8}{l}{\emph{Open-source models}} \\
    Janus-Pro & 7B
      & 0.553 & 0.001 & 0.139 & 0.276 & \textbf{0.365} & 0.267 \\
    BAGEL & 14B
      & 0.769 & 0.244 & 0.173 & 0.367 & \underline{0.251} & 0.361 \\
    HiDream-I1-Full & 17B
      & 0.829 & 0.707 & 0.317 & 0.347 & 0.186 & 0.477 \\
    SD3.5 Large & 8B
      & 0.809 & 0.629 & 0.294 & 0.353 & 0.225 & 0.462 \\
    FLUX.1 [Dev] & 12B
      & 0.786 & 0.523 & 0.253 & 0.368 & 0.238 & 0.434 \\
    FLUX.2-Klein & 9B
      & \underline{0.887} & 0.866 & 0.312 & \textbf{0.442} & 0.156 & 0.532 \\
    Z-Image-Turbo & 6B
      & 0.840 & \textbf{0.994} & 0.298 & 0.368 & 0.139 & 0.528 \\
    Z-Image & 6B
      & 0.881 & \underline{0.987} & 0.280 & 0.387 & 0.194 & 0.546 \\
    Qwen-Image & 20B
      & 0.882 & 0.891 & 0.306 & \underline{0.418} & 0.197 & 0.539 \\
    \midrule
    \textbf{Lens-Turbo} & \textbf{3.8B}
      & 0.884 & 0.972 & \textbf{0.349} & 0.383 & 0.180 & \underline{0.554} \\
    \textbf{Lens} & \textbf{3.8B}
      & \textbf{0.891} & 0.960 & \underline{0.343} & 0.404 & 0.186 & \textbf{0.557} \\
    \bottomrule
  \end{tabular}%
  }
\end{table*}

\subsection{Distillation for T2I Models}

A central challenge of diffusion- and flow-based T2I models is their expensive iterative sampling process. Early acceleration methods mainly improve the inference solver without modifying model parameters. Representative examples include deterministic samplers such as DDIM~\citep{ddim}, as well as high-order ODE solvers such as DPM-Solver~\citep{dpmsolver} and UniPC~\citep{unipc}. Although these training-free methods can reduce the number of function evaluations, generation quality often degrades when the sampling budget becomes extremely small, especially for high-resolution text-conditioned synthesis under strong classifier-free guidance.

\begin{table*}[t]
  \centering
  \caption{Comparison of \textit{Lens} with commercial and open-source models on \textbf{GenEval}.}
  \vspace{1mm}
  \label{tab:main_geneval}
  \small
  \setlength{\tabcolsep}{4.8pt}
  \renewcommand{\arraystretch}{1.12}
  \resizebox{\textwidth}{!}{%
  \begin{tabular}{l|c|ccccccc}
    \toprule
    Model & Size & Single & Two & Count & Color & Position & Attribute & Overall \\
    \midrule
    \multicolumn{9}{l}{\emph{Commercial models}} \\
    Seedream 3.0 & --
      & 0.990 & 0.960 & 0.910 & 0.930 & 0.470 & 0.800 & 0.843 \\
    Seedream 4.0 & --
      & 0.990 & 0.920 & 0.720 & 0.910 & 0.760 & 0.740 & 0.840 \\
    GPT Image 1 [High] & --
      & 0.990 & 0.920 & 0.850 & 0.920 & 0.750 & 0.610 & 0.840 \\
    \midrule
    \multicolumn{9}{l}{\emph{Open-source models}} \\
    Janus-Pro & 7B
      & 0.990 & 0.890 & 0.590 & 0.900 & 0.790 & 0.660 & 0.800 \\
    HiDream-I1-Full & 17B
      & \textbf{1.000} & \textbf{0.980} & 0.790 & 0.910 & 0.600 & 0.720 & 0.833 \\
    SD3.5 Large & 8B
      & 0.980 & 0.890 & 0.730 & 0.830 & 0.340 & 0.470 & 0.710 \\
    FLUX.1 [Dev] & 12B
      & 0.980 & 0.810 & 0.740 & 0.790 & 0.220 & 0.450 & 0.660 \\
    FLUX.2-Klein & 9B
      & 0.931 & 0.957 & 0.828 & 0.915 & 0.718 & 0.740 & 0.848 \\
    Z-Image-Turbo & 6B
      & \textbf{1.000} & 0.950 & 0.770 & 0.890 & 0.650 & 0.680 & 0.823 \\
    Z-Image & 6B
      & \textbf{1.000} & 0.940 & 0.780 & \textbf{0.930} & 0.620 & 0.770 & 0.840 \\
    Qwen-Image & 20B
      & 0.990 & 0.920 & \underline{0.890} & 0.880 & 0.760 & 0.770 & 0.868 \\
    Hunyuan-Image-3.0 & 80B
      & \textbf{1.000} & 0.920 & 0.480 & 0.820 & 0.420 & 0.630 & 0.720 \\
    LongCat-Image & 6B
      & 0.990 & \textbf{0.980} & 0.860 & 0.860 & 0.750 & 0.730 & 0.870 \\
    \midrule
    \textbf{Lens-Turbo} & \textbf{3.8B}
      & \underline{0.997} & 0.947 & \textbf{0.909} & 0.891 & \underline{0.895} & \underline{0.845} & \underline{0.914} \\
    \textbf{Lens} & \textbf{3.8B}
      & 0.994 & \underline{0.970} & \textbf{0.909} & \underline{0.923} & \textbf{0.915} & \textbf{0.868} & \textbf{0.930} \\
    \bottomrule
  \end{tabular}%
  }
\end{table*}

To achieve more aggressive acceleration, another line of work distills a multi-step teacher model into a one-step or few-step student. Progressive distillation~\citep{progressivedistill} iteratively halves the number of sampling steps by matching deterministic teacher trajectories. Consistency models~\citep{consistency} learn a self-consistent mapping from noisy states to clean data, enabling fast generation with few denoising steps. Latent Consistency Models~\citep{latentconsistency} extend this idea to latent-space T2I models, while InstaFlow~\citep{liu2023instaflow} accelerates rectified-flow models by straightening probability-flow trajectories and improving noise-data couplings. These methods demonstrate the potential of trajectory- or consistency-based distillation, but their performance can still degrade under very small step budgets or complex text-conditioned generation.

Recent methods further improve few-step T2I distillation by introducing distribution-level and adversarial supervision. Adversarial Diffusion Distillation~\citep{add} combines score distillation with an adversarial objective to improve perceptual quality. Distribution Matching Distillation~\citep{dmd} directly matches the student distribution to the target data distribution using real and fake score estimates, reducing the reliance on paired teacher trajectories. Its improved variants, including DMD2~\citep{yin2024improved}, decoupled-DMD~\citep{decoupleddmd}, DMD-R~\citep{dmdr}, and SenseFlow~\citep{senseflow}, further enhance training stability, guidance distillation, and distribution alignment for large-scale T2I models. Despite these advances, few-step distillation remains sensitive to teacher guidance, timestep design, fake-score estimation, and adversarial stability.

\begin{table*}[t]
  \centering
  \caption{Comparison of \textit{Lens} with commercial and open-source models on \textbf{LongText (EN)} and \textbf{CVTG}.}
  \vspace{1mm}
  \label{tab:main_longtext_cvtg}
  \small
  \setlength{\tabcolsep}{5pt}
  \renewcommand{\arraystretch}{1.12}
  \resizebox{\textwidth}{!}{%
  \begin{tabular}{l|c|c|ccccccc}
    \toprule
    \multicolumn{1}{l|}{\multirow{2}{*}{\raisebox{-0.6ex}{Model}}}
      & \multicolumn{1}{c|}{\multirow{2}{*}{\raisebox{-0.6ex}{Size}}}
      & \multicolumn{1}{c|}{\multirow{2}{*}{\raisebox{-3.5ex}{\shortstack[c]{LongText\\(EN)}}}}
      & \multicolumn{7}{c}{CVTG} \\
    \cmidrule{4-10}
      & & & 2R & 3R & 4R & 5R & Avg. & NED & CLIP \\
    \midrule
    \multicolumn{10}{l}{\emph{Commercial models}} \\
    Kolors 2.0 & --
      & 0.258
      & -- & -- & -- & -- & -- & -- & -- \\
    Seedream 3.0 & --
      & 0.896
      & 0.628 & 0.596 & 0.604 & 0.561 & 0.592 & 0.854 & 0.782 \\
    Seedream 4.0 & --
      & 0.921
      & 0.890 & 0.915 & 0.899 & 0.887 & 0.892 & 0.951 & 0.785 \\
    GPT Image 1 [High] & --
      & 0.956
      & 0.878 & 0.866 & 0.873 & 0.822 & 0.857 & 0.948 & 0.798 \\
    Nano Banana 2.0 & --
      & 0.981
      & -- & -- & -- & -- & -- & -- & -- \\
    \midrule
    \multicolumn{10}{l}{\emph{Open-source models}} \\
    Janus-Pro & 7B
      & 0.019
      & -- & -- & -- & -- & -- & -- & -- \\
    BAGEL & 14B
      & 0.373
      & -- & -- & -- & -- & -- & -- & -- \\
    HiDream-I1-Full & 17B
      & 0.543
      & -- & -- & -- & -- & -- & -- & -- \\
    SD3.5 Large & 8B
      & --
      & 0.729 & 0.683 & 0.657 & 0.594 & 0.655 & 0.847 & 0.780 \\
    FLUX.1 [Dev] & 12B
      & 0.607
      & 0.609 & 0.553 & 0.466 & 0.432 & 0.496 & 0.688 & 0.740 \\
    FLUX.2-Klein & 9B
      & 0.864
      & -- & -- & -- & -- & -- & -- & -- \\
    Z-Image-Turbo & 6B
      & 0.917
      & 0.887 & 0.866 & 0.863 & 0.835 & 0.859 & 0.928 & 0.805 \\
    Z-Image & 6B
      & 0.935
      & \underline{0.901} & 0.872 & 0.865 & \underline{0.851} & 0.867 & 0.937 & 0.797 \\
    Qwen-Image & 20B
      & \textbf{0.943}
      & 0.837 & 0.836 & 0.831 & 0.816 & 0.829 & 0.930 & 0.806 \\
    Hunyuan-Image-3.0 & 80B
      & --
      & 0.830 & 0.764 & 0.738 & 0.728 & 0.765 & 0.877 & 0.812 \\
    LongCat-Image & 6B
      & --
      & \textbf{0.913} & 0.874 & 0.856 & 0.831 & 0.866 & 0.936 & 0.786 \\
    \midrule
    \textbf{Lens-Turbo} & \textbf{3.8B}
      & 0.927
      & 0.893 & \textbf{0.892} & \textbf{0.894} & \textbf{0.878} & \textbf{0.889} & \textbf{0.965} & \textbf{0.815} \\
    \textbf{Lens} & \textbf{3.8B}
      & \underline{0.937}
      & 0.897 & \underline{0.881} & \underline{0.872} & 0.827 & \underline{0.869} & \underline{0.951} & \underline{0.814} \\
    \bottomrule
  \end{tabular}%
  }
\end{table*}

\subsection{VAE}
Variational Autoencoders (VAEs)~\citep{kingma2013vae} are widely used as image tokenizers in diffusion-based generation models, serving as the bridge between pixel space and latent representations. Conventional tokenizers are typically trained with reconstruction-oriented objectives, aiming to preserve as much pixel-level information as possible. However, recent studies suggest that latents optimized purely for reconstruction are not necessarily optimal for generative modeling. Such latents may be semantically entangled, difficult for diffusion models to learn, and can slow down convergence during training. \textit{Reconstruction vs. Generation}~\citep{liu2024reconstruction} formally analyzes this conflict in latent diffusion models, while \textit{Both Semantics and Reconstruction Matter}~\citep{yang2024both} empirically shows that excessive reconstruction pressure tends to prioritize low-level details over semantic structure, which can hurt text-to-image generation and editing performance.

Motivated by this mismatch, recent work has explored generation-friendly tokenizers that better align the latent space with downstream generative objectives. REPA-E~\citep{leng2025repae} aligns encoder representations with diffusion Transformer features, thereby easing optimization for the generative model. Unified Latents~\citep{hu2024unifiedlatents} jointly optimizes reconstruction and generation objectives during tokenizer training, reducing the gap between tokenizer learning and generative modeling by design. Latent Forcing~\citep{tu2024latentforcing} further improves generation by imposing latent-level constraints and reorganizing the diffusion trajectory. These methods demonstrate that the quality of a tokenizer should be evaluated not only by reconstruction fidelity, but also by how effectively its latent space supports generative learning.

Another complementary direction improves tokenizer capacity, structure, and semantic awareness. VTP~\citep{vtp} incorporates visual understanding tasks into tokenizer pretraining, encouraging the learned latents to encode richer semantic information. MagViT-v2~\citep{yu2023magvitv2}, VAR~\citep{tian2024visual}, and TiTok~\citep{yu2024titok} explore masked, multiscale, discrete, or sequential latent representations to improve generative learnability and scalability. In addition, distillation-based tokenizers leverage pretrained vision models such as CLIP~\citep{radford2021clip} and DINO~\citep{cijosevic2021dino} to inject semantic structure into the latent space. Overall, these studies indicate a broader shift from reconstruction-optimal tokenizers toward generation-oriented tokenizers, where semantic organization, learnability, and compatibility with downstream generative models are treated as central design goals.

\section{More Results}
\label{sec:detailed-results}

\subsection{Detailed Benchmark Results}
\label{sec:benchmark-details}

Tables~\ref{tab:main_oneig},~\ref{tab:main_geneval} and~\ref{tab:main_longtext_cvtg} show the comparison of \textit{Lens} against state-of-the-art models on \textbf{OneIG (EN)}, \textbf{GenEval}, \textbf{LongText (EN)} and \textbf{CVTG}, respectively.

\begin{table*}[t]
  \centering
  \caption{Comparison of \textit{Lens} variants with different reasoners. All models are non-distilled versions using 20-step denoising and differ only in the choice of reasoner. We also compare with Qwen-Image equipped with a GPT-5.5 reasoner, using the same system prompt optimized by our training-free prompt search strategy. The results show that this strategy generalizes to other T2I models.}
  \vspace{1mm}
  \label{tab:different reasoners}
\begin{tabular}{l|c|c|c|c|ccc}
  \toprule
  \multicolumn{1}{l|}{\multirow{2}{*}{\raisebox{-0.6ex}{Model}}}
    & \multicolumn{1}{c|}{\multirow{2}{*}{\raisebox{-0.6ex}{Size}}}
    & \multicolumn{1}{c|}{\multirow{2}{*}{\raisebox{-3ex}{\shortstack[c]{OneIG\\(EN)}}}}
    & \multicolumn{1}{c|}{\multirow{2}{*}{\raisebox{-0.6ex}{GenEval}}}
    & \multicolumn{1}{c|}{\multirow{2}{*}{\raisebox{-3ex}{\shortstack[c]{LongText\\(EN)}}}}
    & \multicolumn{3}{c}{CVTG} \\
  \cmidrule(l){6-8}
   & & & & & Avg. & NED & CLIP \\
  \midrule
  
  Lens w/o reasoner
    & 3.8B & 0.532 & 0.843 & 0.893 & 0.849 & 0.933 & 0.796\\
  \midrule
  Qwen-Image w/ GPT-5.5 & 20B & \textbf{0.567} & 0.926 & \textbf{0.962} & \textbf{0.891} & 0.947 & 0.787 \\
  Lens w/ GPT-5.5
   & 3.8B & 0.557 & \textbf{0.930} & 0.937 & 0.869 & 0.951 & 0.814 \\
  Lens w/ GPT-OSS-20BA3B
    & 3.8B & 0.559 & 0.874 & 0.924 & 0.888 & \textbf{0.958} & 0.821 \\
  Lens w/ Qwen3-0.6B & 3.8B & 0.522 & 0.820 & 0.866 & 0.865 & 0.943 & 0.800 \\
  Lens w/ Qwen3-1.7B & 3.8B & 0.542 & 0.875 & 0.912 & 0.864 & 0.942 & 0.801 \\
  Lens w/ Qwen3-4B & 3.8B & 0.546 & 0.871 & 0.922 & 0.883 & 0.953 & \textbf{0.823} \\
  \bottomrule
\end{tabular}%

%}
\vspace{-2mm}
\end{table*}

\subsection{Various Reasoners}
\label{sec:various reasoners}

In Table~\ref{tab:different reasoners}, we compare \textit{Lens} variants equipped with different reasoners, including no reasoner, GPT-5.5, GPT-OSS-20BA3B, and Qwen3 models of different sizes (0.6B, 1.7B, and 4B).

\section{Visualization}
\label{sec:visualization-appendix}

\subsection{Prompt-rubric Visualization}
\label{vis:rubric}
We provide several prompt-rubric examples from our \textit{Lens-RL-8K} prompt set below.

\begin{tcolorbox}[
  breakable,
  enhanced,
  colback=white,
  colframe=black!70,
  boxrule=0.8pt,
  arc=2pt,
  title={Prompt-rubric Examples},
  colbacktitle=black!75,
  coltitle=white,
  fonttitle=\bfseries,
  left=8pt,right=8pt,top=8pt,bottom=8pt
]
\footnotesize
\begin{description}
\item[\textcolor{blue}{Example 1}]
\item[Prompt:] "On a sunny playground, a school-age child stands in the left foreground, gripping 1 bright red kickball, shot at waist level with crisp midday shadows; behind, 5 empty swings arc gently. A metal fence sign reads 'PLAYGROUND RULES – WALK PLEASE.' Blue sneakers and dusty knees emphasize lively texture."
\item[Rubrics:] [["Object Count Consistency (Kickball)", "Verify that exactly one bright red kickball is present in the image."], ["Object Count Consistency (Empty Swings)", "Verify that exactly five empty swings are present, as specified in the prompt."], ["Object Count Consistency (School-age Child)", "Verify that exactly one school-age child is shown, as specified in the prompt."], ["Object Placement and Spatial Reasoning (Child Position)", "Ensure the child is positioned in the left foreground of the playground scene."], ["Object Placement and Spatial Reasoning", "Ensure the child is positioned in the left foreground with the swings visible behind at an appropriate distance and angle."], ["Action Accuracy (Grip Ball)", "Ensure the child is shown gripping the kickball at waist level."], ["OCR Alignment (Fence Sign)", "Ensure the metal fence sign text matches 'PLAYGROUND RULES – WALK PLEASE.'"], ["Attribute Accuracy (Kickball \& Child Details)", "Verify the kickball is bright red and the child wears blue sneakers with dusty knees, and that the scene has crisp midday shadows."], ["Attribute Accuracy (Kickball Color)", "Verify that the kickball is bright red as specified."], ["Attribute Accuracy (Sneakers and Knees Texture)", "Verify the child is wearing blue sneakers and the knees appear dusty with a textured look."], ["Structural Integrity (Overall)", "Verify that the entire image is structurally coherent and physically plausible, with no missing, duplicated, disconnected, distorted, floating, merged, or incorrectly placed parts; all subjects, objects, perspective relationships, and spatial connections should appear complete, aligned, and naturally integrated into the scene."]]

\item[\textcolor{blue}{Example 2}]
\item[Prompt:] "A historic clock tower of warm ochre sandstone stands centered in the frame, shot low-angle with a 35mm lens at golden hour, its green-patinated copper roof and ivory clock face with black numerals catching amber light, surrounding plaza of gray cobblestones and red banners fading into soft bokeh."
\item[Rubrics:] [["Object Placement and Spatial Reasoning", "Ensure the clock tower is centered in the frame and shot from a low-angle perspective as specified."], ["Object Count Consistency (Clock Tower)", "Verify that only one clock tower is depicted, as specified."], ["Attribute Accuracy (Sandstone Color)", "Verify the tower appears in a warm ochre sandstone hue without color discrepancies."], ["Attribute Accuracy (Roof Patina)", "Verify the roof displays a green-patinated copper appearance."], ["Attribute Accuracy (Clock Face \& Numerals)", "Verify the clock face is ivory with clearly rendered black numerals."], ["Attribute Accuracy (Lighting/Golden Hour)", "Verify the scene exhibits amber light characteristic of golden hour."], ["Attribute Accuracy (Cobblestone Plaza)", "Ensure the surrounding plaza is made of gray cobblestones."], ["Attribute Accuracy (Banner Color and Bokeh)", "Verify the red banners are present and fade into a soft bokeh."], ["Attribute Accuracy (Lighting and Bokeh)", "Verify golden hour amber light casts on the scene and the background exhibits soft bokeh."], ["Attribute Accuracy (Plaza and Banners)", "Confirm the plaza is paved with gray cobblestones and red banners fade into the soft background."], ["Structural Integrity (Overall)", "Verify that the entire image is structurally coherent and physically plausible, with no missing, duplicated, disconnected, distorted, floating, merged, or incorrectly placed parts; all subjects, objects, perspective relationships, and spatial connections should appear complete, aligned, and naturally integrated into the scene."]]

\item[\textcolor{blue}{Example 3}]
\item[Prompt:] "An old library reading room under warm lamps, wood paneling and quiet dust. Shot from a mezzanine, the central aisle runs centrally between four long oak tables, guiding the eye to a distant circulation desk. Readers sit on both sides, their chairs tucked in, emphasizing the aisle’s dividing relationship and middle placement."
\item[Rubrics:] [["Object Placement and Spatial Reasoning", "Verify the central aisle runs centrally between the four tables and leads the eye to the distant circulation desk."], ["Object Count Consistency (Oak Tables)", "Verify exactly four long oak tables are present as specified in the prompt."], ["Attribute Accuracy (Warm Lamps)", "Verify the scene is illuminated by warm lamp lighting creating a cozy, amber glow."], ["Attribute Accuracy (Wood Paneling)", "Verify the walls and surroundings feature authentic wood paneling consistent with an old library."], ["Attribute Accuracy (Dusty Atmosphere)", "Verify a subtle dustiness is visible in the air and on surfaces to convey quiet age."], ["Object Count Consistency (Circulation Desk)", "Verify that one circulation desk appears at the far end of the aisle."], ["Attribute Accuracy (Lamp Warmth)", "Verify that the lamps cast a warm, inviting glow consistent with the prompt."], ["Attribute Accuracy (Oak Tables)", "Verify that there are four long oak tables with appropriate grain and hue."], ["Object Placement and Spatial Reasoning (Central Aisle)", "Verify that the aisle runs centrally between the tables leading to a distant circulation desk, with readers seated on both sides and chairs tucked in."], ["Attribute Accuracy (Dust Atmosphere)", "Ensure subtle dust particles are visible in the light to convey a quiet, dusty ambiance."], ["Structural Integrity (Overall)", "Verify that the entire image is structurally coherent and physically plausible, with no missing, duplicated, disconnected, distorted, floating, merged, or incorrectly placed parts; all subjects, objects, perspective relationships, and spatial connections should appear complete, aligned, and naturally integrated into the scene."]]

\item[\textcolor{blue}{Example 4}]
\item[Prompt:] "A telephoto close-up of dense black-and-white fur on a panda's shoulder occupying the right foreground, morning zoo light, dewy green bamboo blurred behind; on the left, a brushed-steel placard reads 'GIANT PANDA — Ailuropoda melanoleuca', center-left placement, crisp engraved letters catching highlights."
\item[Rubrics:] [["Object Placement and Spatial Reasoning (Panda Shoulder Right Foreground)", "Verify that the panda’s shoulder is positioned in the right foreground of the image."], ["Object Placement and Spatial Reasoning (Placard Center-Left)", "Verify that the placard is positioned in the center-left of the image."], ["Attribute Accuracy (Panda Fur Texture)", "Verify the fur appears dense, black-and-white with detailed texture under morning zoo light."], ["Attribute Accuracy (Bamboo Background Blur)", "Verify the green bamboo appears dewy and softly blurred behind the panda."], ["Attribute Accuracy (Placard Material and Finish)", "Check the placard has a brushed-steel look with crisp engraved lettering catching highlights."], ["OCR Alignment (Placard Text)", "Ensure the engraved text ‘GIANT PANDA — Ailuropoda melanoleuca’ is spelled, punctuated, and cased exactly as in the prompt."], ["OCR Alignment", "Ensure the placard text reads exactly 'GIANT PANDA — Ailuropoda melanoleuca' in crisp engraved letters."], ["Attribute Accuracy (Background Blur and Color)", "Confirm the bamboo backdrop is blurred with a dewy green hue indicating shallow depth of field."], ["Object Count Consistency (Placard)", "Verify that exactly one placard is present."], ["Object Placement and Spatial Reasoning", "Verify the panda's shoulder occupies the right foreground and the placard is placed center-left with blurred bamboo behind."], ["Structural Integrity (Overall)", "Verify that the entire image is structurally coherent and physically plausible, with no missing, duplicated, disconnected, distorted, floating, merged, or incorrectly placed parts; all subjects, objects, perspective relationships, and spatial connections should appear complete, aligned, and naturally integrated into the scene."]]

\end{description}
\end{tcolorbox}

\subsection{Lens-800M Training Data Visualization}
\label{sec:lens-800M visualization}
\begin{figure}[!t]
    \centering
    \includegraphics[width=0.99\textwidth]{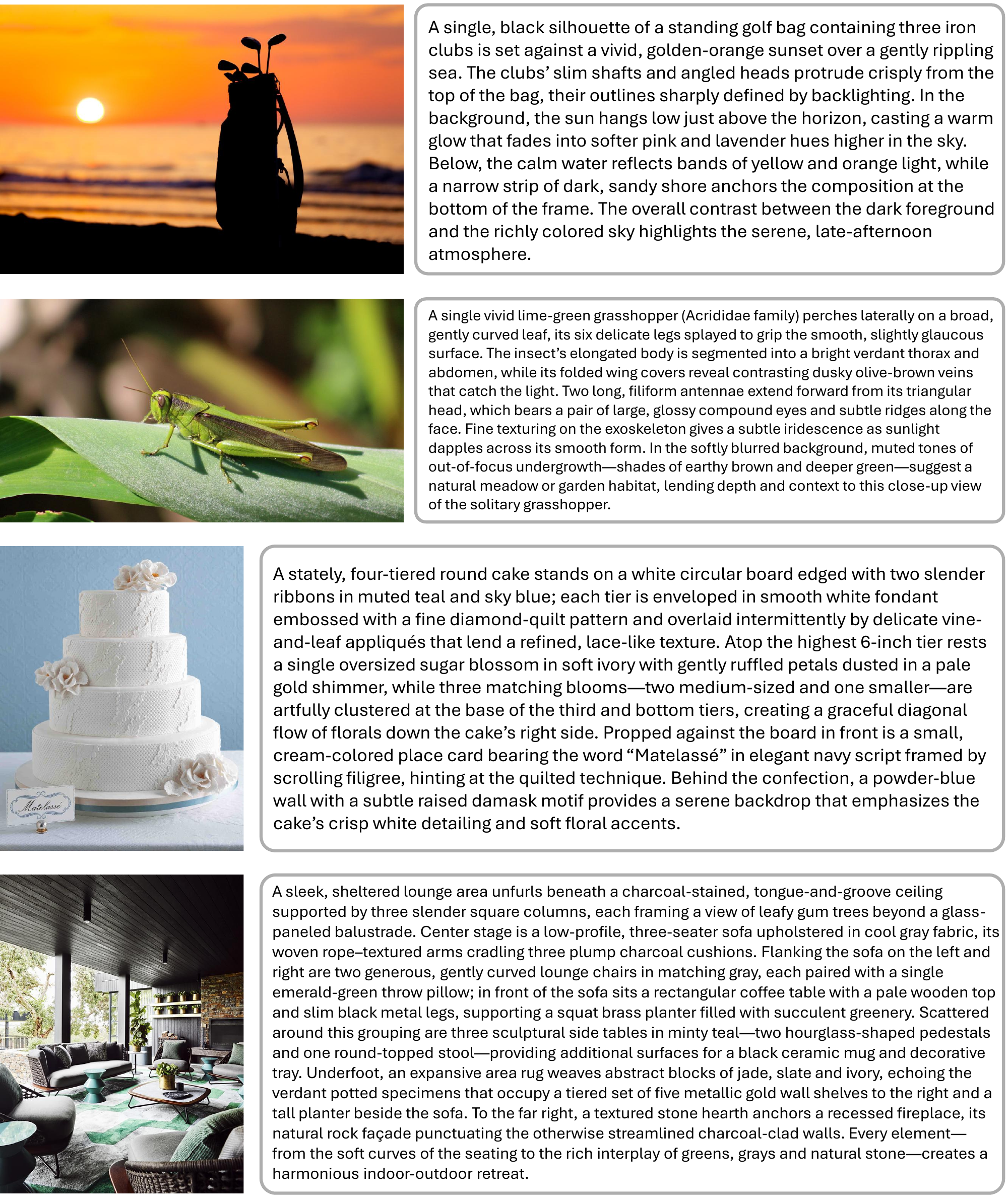}
    \caption{
  Each row shows a densely captioned image-text pair from our \textit{Lens-800M} training set.
}
    \label{fig:vis-training-samples}
\end{figure}

In Figure~\ref{fig:vis-training-samples}, we show several examples from our \textit{Lens-800M} training set, where each training sample is a densely captioned image-text pair.

\section{Implementation Details}
\label{sec:implementation details}

\subsection{RL Details}
\label{sec:RL-Details}

\noindent\textbf{Preliminary of DiffusionNFT.}
Unlike traditional reinforcement learning (RL) methods that optimize generative models through policy-gradient objectives~\citep{liu2025flow,xue2025dancegrpo}, Diffusion Negative-aware Finetuning (DiffusionNFT)~\citep{zheng2025diffusionnft} performs reward-based policy optimization directly within the forward diffusion process using the flow-matching objective. Its key idea is to use the reward signal to distinguish desirable and undesirable generation directions. Specifically, DiffusionNFT trains the flow-matching model (FMM) to learn not only a \emph{positive} velocity $v^{+}(x_t,c,t)$ that moves samples toward high-reward generations, but also a \emph{negative} velocity $v^{-}(x_t,c,t)$ that represents directions the model should avoid.

The core policy-optimization loss is defined as:
\begin{equation*}
\label{eq:diffusion_nft_loss}
\mathcal{L}(\theta) =
\mathbb{E}_{c,\pi^{\mathrm{old}}(x_0 \mid c),t}
\Big[
r \, \| v_\theta^{+}(x_t,c,t) - v \|_2^2
+
(1-r) \, \| v_\theta^{-}(x_t,c,t) - v \|_2^2
\Big],
\end{equation*}
where $v$ denotes the target velocity field, and $r \in [0,1]$ is the normalized reward interpreted as the optimality probability of the generated sample. The positive and negative velocities, $v_\theta^{+}$ and $v_\theta^{-}$, are implicitly defined as linear combinations of the old policy $v^{\mathrm{old}}$ and the current policy $v_\theta$, controlled by a weighting coefficient $\beta$:
\begin{equation*}
\label{eq:diffusion_nft_velocities}
\begin{aligned}
v_\theta^{+}(x_t,c,t) &:=
(1-\beta)\,v^{\mathrm{old}}(x_t,c,t)
+
\beta\,v_\theta(x_t,c,t), \\
v_\theta^{-}(x_t,c,t) &:=
(1+\beta)\,v^{\mathrm{old}}(x_t,c,t)
-
\beta\,v_\theta(x_t,c,t).
\end{aligned}
\end{equation*}
Intuitively, high-reward samples assign larger weight to the positive velocity term, encouraging the current policy to move toward desirable directions. Conversely, low-reward samples emphasize the negative velocity term, pushing the model away from undesirable generation trajectories.

Since unconstrained raw rewards $r^{\mathrm{raw}}(x_0,c)$ may vary substantially in scale and distribution across prompts, DiffusionNFT converts them into a bounded optimality probability $r \in [0,1]$:
\begin{equation*}
\label{eq:diffusion_nft_reward}
r(x_0,c) :=
\frac{1}{2}
+
\frac{1}{2}
\operatorname{clip}
\left[
\frac{
r^{\mathrm{raw}}(x_0,c)
-
\mathbb{E}_{\pi^{\mathrm{old}}(\cdot \mid c)}
\big[r^{\mathrm{raw}}(x_0,c)\big]
}{Z_c},
-1, 1
\right],
\end{equation*}
where $Z_c > 0$ is a normalization factor, typically set to the global standard deviation of rewards. This normalization stabilizes training by mapping raw reward differences into a bounded probability-like signal, enabling the model to balance positive-direction learning and negative-direction avoidance during finetuning.

\noindent\textbf{Optimization Details.}
Starting from the pre-trained checkpoint, we perform reinforcement learning (RL) for 180 steps on the \textit{Lens-RL-8K} dataset using 64 NVIDIA A100 80GB GPUs. To maintain high-fidelity generation across diverse image layouts, we use mixed-resolution training buckets with a fixed base area of $1024^2$. Specifically, we consider nine aspect ratios: $736\times1472$, $768\times1376$, $832\times1248$, $864\times1152$, $1024\times1024$, $1152\times864$, $1248\times832$, $1376\times768$, and $1472\times736$.

We fine-tune the diffusion Transformer using Low-Rank Adaptation (LoRA)~\citep{hu2022lora}, with rank $r{=}64$ and scaling factor $\alpha{=}128$. For the RL objective, we follow DiffusionNFT~\citep{zheng2025diffusionnft} and adopt a group-based optimization strategy, using 48 groups per epoch and a group size of 24. For each clean image in the dataset, we apply forward noising and compute the training loss at the corresponding sampling timesteps. We use a second-order ODE sampler for trajectory collection and incorporate adaptive time weighting to stabilize optimization.

To reduce reward hacking and preserve generation diversity, we apply a KL-divergence penalty with coefficient $\beta_{\text{KL}}{=}1\times10^{-4}$. We optimize the model using AdamW with $\beta_1{=}0.9$, $\beta_2{=}0.999$, $\epsilon{=}10^{-8}$, and weight decay $1\times10^{-4}$. Following DiffusionNFT, we set the learning rate to $3\times10^{-4}$, use $\beta{=}1$, and define the adaptive coefficient as $\eta_i=\min(0.001,0.5)$ for training stabilization.

\subsection{Distillation Details}
\label{sec:Distill-Details}

\noindent\textbf{Preliminary of DMD.}
Distribution Matching Distillation (DMD)~\citep{dmd} distills a multi-step diffusion model into a few-step generator by matching the student distribution to the target data distribution. Given a condition $c$ and noise $z$, let $x_\theta=G_\theta(z,c)$ be the student output. DMD minimizes a reverse-KL-style objective:
\begin{equation*}
\mathcal{L}_{\mathrm{DMD}}
=
D_{\mathrm{KL}}
\left(
p_\theta(x_0 \mid c)
\;\|\;
p_{\mathcal{D}}(x_0 \mid c)
\right),
\end{equation*}
where $p_\theta$ is induced by the student and $p_{\mathcal{D}}$ denotes the data distribution. Since the density is intractable, DMD optimizes this objective through an approximate score-based gradient. Given
\begin{equation*}
x_t=\alpha_t x_\theta+\sigma_t\epsilon,
\quad
\epsilon\sim\mathcal{N}(0,I),
\end{equation*}
the gradient can be estimated as
\begin{equation*}
\nabla_\theta \mathcal{L}_{\mathrm{DMD}}
\approx
\mathbb{E}_{c,z,t,\epsilon}
\left[
\left(
s_\phi(x_t,c,t)
-
s_\psi(x_t,c,t)
\right)
\frac{\partial x_\theta}{\partial \theta}
\right],
\end{equation*}
where $s_\psi$ is provided by the frozen teacher and $s_\phi$ is produced by a fake score model trained on student-generated samples. The teacher score pulls samples toward the target distribution, while the fake score counteracts over-concentration of the student distribution.

\noindent\textbf{Optimization Details.}
We curate a 100K image-caption subset from \textit{Lens-800M} based on aesthetic and other quality-related metrics, while balancing major generation scenarios such as portraits, landscapes, visual content, artistic styles, and text-rich images. The captions are used as text conditions for the DMD objectives, and the paired images provide real samples for discriminator training. Following pre-training and RL, we use mixed-resolution bucketed loading for both student backward simulation and discriminator training to preserve generation capability across different layouts.

We initialize the student $G_\theta$ and the fake score model $s_\phi$ from the RL-aligned checkpoint and train them with full-parameter optimization, while keeping the teacher $s_\psi$ frozen. Following decoupled-DMD~\citep{decoupleddmd}, we decompose the student objective into a CFG-augmented term $\mathcal{L}_{\mathrm{CA}}$ and a distribution-matching term $\mathcal{L}_{\mathrm{DM}}$. The former distills the guidance effect into the student, while the latter performs distribution matching through the fake score model.

To complement the score-based objective with direct real-data supervision, we adopt an adversarial branch as in DMD2~\citep{dmd2}. Let
\begin{equation*}
q_t(x) \sim \mathcal{N}(\alpha_t x, \sigma_t^2 I)
\end{equation*}
denote a sample from the forward noising process. The discriminator compares noised real samples $q_t(x)$ and noised student samples $q_t(x_\theta)$ at the same diffusion time $t$, where $x\sim p_{\mathcal{D}}(\cdot\mid c)$ and $x_\theta=G_\theta(z,c)$. Different from DMD2, which uses intermediate features from the fake score model, our discriminator operates on features extracted by the frozen teacher. Let
\begin{equation*}
d_\eta(x_t,c,t)=D_\eta(h_\psi(x_t,c,t))
\end{equation*}
denote the discriminator logit, where $h_\psi$ is the frozen teacher feature extractor. We use the logistic loss $\ell(r)=\log(1+\exp(-r))$. The discriminator objective is
\begin{equation*}
\begin{aligned}
\mathcal{L}_{\eta}
=&\;
\mathbb{E}_{x,c,t}
\left[
\ell\!\left(d_{\eta}(q_t(x),c,t)\right)
\right]
+
\mathbb{E}_{z,c,t}
\left[
\ell\!\left(-d_{\eta}(q_t(x_{\theta}),c,t)\right)
\right] \\
&+
\frac{\gamma}{2}
\mathbb{E}_{x,c,t,\epsilon}
\left[
\left\|
d_{\eta}(q_t(x),c,t)
-
d_{\eta}(\bar{x}_t,c,t)
\right\|_{2}^{2}
\right],
\end{aligned}
\end{equation*}
where $\bar{x}_t=q_t(x)+\alpha\epsilon$ is a small perturbation of the noised real sample for approximating the R1 penalty. We set $\gamma=1.0$ and $\alpha=0.1$ in practice. The generator-side adversarial loss is
\begin{equation*}
\mathcal{L}_{G}
=
\mathbb{E}_{z,c,t}
\left[
\ell\!\left(d_\eta(q_t(x_\theta),c,t)\right)
\right].
\end{equation*}

Thus the student $G_\theta$ is optimized with
\begin{equation*}
\mathcal{L}_{\theta}
=
\lambda_d
\left(
\mathcal{L}_{\mathrm{DM}}
+
\mathcal{L}_{\mathrm{CA}}
\right)
+
\lambda_g
\mathcal{L}_{G},
\end{equation*}
where $\lambda_d=0.1$ and $\lambda_g=0.001$. The fake score model $s_\phi$ is trained with a velocity-matching objective on noised student samples:
\begin{equation*}
\mathcal{L}_{\phi}
=
\mathbb{E}_{c,z,t}
\left[
\left\|
v_\phi(q_t(x_\theta),c,t)-u_t
\right\|_2^2
\right],
\end{equation*}
where $x_\theta=G_\theta(z,c)$ and $u_t$ denotes the flow-matching target associated with the same forward-noised sample $q_t(x_\theta)$.

Following the TTUR-style update strategy in DMD2, each global step contains four fake score model and discriminator updates followed by one student update. After each student update, we further adopt the IDA strategy from SenseFlow~\citep{senseflow}, updating the fake score model toward the student via $\phi \leftarrow (1-\mu) \cdot \phi+\mu \cdot \theta$ where $\mu=0.03$.

Putting everything together, the student is trained for 4-step generation, with guidance scale $5.0$ for $\mathcal{L_{\mathrm{CA}}}$ during distillation. We train on 8 NVIDIA A100 80GB GPUs with per-GPU batch size 4. We use AdamW with learning rate $5\times10^{-7}$ for the student and fake score model, and $1\times10^{-4}$ for the discriminator, with $\beta_1=0.0$ and $\beta_2=0.9$. Training is conducted for up to 1K global steps.

\section{Prompt}
\label{sec:all-prompts}

\subsection{Prompt for Image Captioning}
\label{sec:prompt-caption}

We use the following prompt to generate a dense caption for each image in the \textit{Lens-800M} pre-training dataset.

\begin{tcolorbox}[
  breakable,
  enhanced,
  colback=white,
  colframe=black!70,
  boxrule=0.8pt,
  arc=2pt,
  title={Prompt for Image Captioning},
  colbacktitle=black!75,
  coltitle=white,
  fonttitle=\bfseries,
  left=8pt,right=8pt,top=8pt,bottom=8pt
]
\footnotesize
Generate detailed caption for the image with 1 paragraph with less than 500 words. Include most salient objects with their attributes (e.g., color/texture/posture) and relationship/interaction, possibly the count number of each object type, and the background. If any text is present in the image (e.g., signs, labels), include the text content in its original language, otherwise MUST NOT mention whether it contains text like description "no visible text". Make sure the caption is comprehensive and accurate enough to reconstruct the image visually.  If the image contains well known contents (e.g., landmark, celebrity, characters, popular dishes), please include such world knowledge in the caption, rather than just give common description.
\end{tcolorbox}

\subsection{Prompt for Lens-RL-8K Dataset Construction}
\label{sec:prompt-lens-RL-8K}
We use the following prompt to generate image-generation prompts during the construction of the \textit{Lens-RL-8K} dataset.

\begin{tcolorbox}[
  breakable,
  enhanced,
  colback=white,
  colframe=black!70,
  boxrule=0.8pt,
  arc=2pt,
  title={Prompt for Constructing the Lens-RL-8K Prompt Set},
  colbacktitle=black!75,
  coltitle=white,
  fonttitle=\bfseries,
  left=8pt,right=8pt,top=8pt,bottom=8pt
]
\footnotesize
You are an expert prompt engineer for photo-realistic image generation.

Given an entity (a concept, object, person type, scene, etc.) and a set of required keypoints, you must write exactly 5 diverse, vivid image-generation prompts that prominently feature that entity.

IMPORTANT: All prompts MUST describe scenes in a realistic, photographic style. Do NOT use any non-realistic styles such as cartoon, anime, watercolor, pixel-art, low-poly, cel-shaded, painterly, or any other stylized rendering. Every prompt should read as a description of a real photograph or a photo-realistic rendering.

\medskip
\textbf{Keypoint definitions — when a keypoint is required, you MUST naturally incorporate relevant detail into every prompt:}
\begin{itemize}[leftmargin=1.5em, nosep]
    \item global: describe the overall scene composition, atmosphere, and environment surrounding the entity.
    \item counting: include specific quantities of objects or subjects (e.g., "three apples", "a pair of shoes").
    \item position: specify spatial placement or pose of the entity (e.g., "in the center", "on the left side", "foreground vs background").
    \item attribute: describe distinctive physical attributes (e.g., shape, size, material, texture, condition).
    \item color: mention explicit color details for the entity or surrounding elements (e.g., "deep crimson", "pale blue").
    \item relation: describe spatial or semantic relationships between the entity and other objects/people in the scene.
    \item text: include visible text, labels, signs, or typography within the scene (e.g., a storefront sign, a book title, a label on a bottle).
\end{itemize}

\textbf{Rules:}
\begin{itemize}[leftmargin=1.5em, nosep]
    \item Each prompt should be 30-60 words, written as a single descriptive paragraph (no line breaks).
    \item Prompts must be concrete and cinematic: specify subject details, setting, lighting, camera angle, mood, and realistic texture cues.
    \item The 5 prompts should vary in setting, lighting, composition, and mood — avoid repetition.
    \item The entity must be the clear focal subject of every prompt.
    \item ALL required keypoints must be naturally woven into EVERY prompt — do not ignore any.
    \item Every prompt must use a realistic, photographic style — no stylized, illustrated, or non-photographic aesthetics.
    \item Return ONLY a valid JSON object with exactly these keys: "prompt-1", "prompt-2", "prompt-3", "prompt-4", "prompt-5". No extra keys, no markdown fences, no explanation.
\end{itemize}

\textbf{Style and length reference (ONLY for tone, structure, and approximate length — do NOT copy or imitate the specific content, subjects, or settings from these examples; your prompts must be entirely about the given entity):}
\begin{itemize}[leftmargin=1.5em, nosep]
    \item A close-up portrait of a young woman with clear skin, dark almond-shaped eyes, and long black hair, facing the camera directly, neutral expression, soft daylight from a nearby window, natural skin texture, realistic eyelashes, subtle lip color, shallow depth of field, photo-realistic.
    \item A weathered brick wall with deep terracotta tones and chalky mortar lines, late afternoon side-light raking across the rough surface, revealing fine granular texture, hairline cracks, and patches of pale efflorescence, sharp macro detail, realistic masonry photography.
    \item Exactly three ripe lemons arranged in a loose triangle on a dark slate countertop, overhead soft-box lighting, each lemon casting a faint shadow, visible pore texture on bright yellow skin, shallow depth of field blurring the wooden cutting board behind, photo-realistic still life.
    \item A gray tabby cat sitting to the left of a tall ceramic vase on a sunlit windowsill, the cat's tail curling around the vase base, warm morning light from the right, lace curtain casting soft shadow patterns, realistic fur detail and glossy ceramic reflection.
    \item A vintage neon sign reading 'OPEN 24 HRS' mounted above a rain-soaked diner entrance at night, buzzing warm amber and cool blue tubes, wet pavement reflecting the glow, peeling paint on the door frame, realistic glass tube filament detail, urban night photography.
\end{itemize}
\end{tcolorbox}

\subsection{Prompt for Rubric Generation}
\label{sec:prompt-rubric}

We adopt the following system prompt to generate rubrics for each prompt in our RL prompt set, \textit{Lens-RL-8K}.

\begin{tcolorbox}[
  breakable,
  enhanced,
  colback=white,
  colframe=black!70,
  boxrule=0.8pt,
  arc=2pt,
  title={Prompt for Rubric Generation},
  colbacktitle=black!75,
  coltitle=white,
  fonttitle=\bfseries,
  left=8pt,right=8pt,top=8pt,bottom=8pt
]
\footnotesize
Please output up to 10 rubrics in a single JSON object.\\
Each key should be a \textbf{short rubric name} (e.g., "Object Count Consistency"), and each value should be a \textbf{brief explanation}.\\
Valid example:\\
\{"Object Count Consistency": "Verify that exactly one cat is shown.", "OCR Alignment": "Ensure visible text matches prompt description."\}\\
Invalid example (CSV format):\\
"Object Count Consistency","Verify that exactly one cat is shown."\\
Invalid examples (DO NOT DO THESE):\\
- \{"Object counting": "..."\} \hspace{1em} \# INVALID renamed base key\\
- \{"Count Object": "..."\} \hspace{1em} \# INVALID reordered/changed\\
- \{"Object  Count  Consistency": "..."\} \# INVALID spacing changed\\
Strictly follow this format:\\
- A single JSON object (not a list)\\
- Keys and values wrapped in double quotes\\
- do not change the key name\\
- No trailing commas or comments\\
- No line breaks within keys or values\\
Your output must be valid JSON parsable with Python's "json.loads()".
\end{tcolorbox}

\subsection{Prompt for RL Reward Generation}
\label{sec:prompt-reward}
Given a prompt and its corresponding rubrics from our \textit{Lens-RL-8K} set, we first generate an image using the current policy model. We then use GPT-4.1-mini as the reward function, with the following system prompt, to evaluate whether the generated image satisfies the rubrics and to produce the reward signal.

\begin{tcolorbox}[
  breakable,
  enhanced,
  colback=white,
  colframe=black!70,
  boxrule=0.8pt,
  arc=2pt,
  title={Prompt for RL Reward Generation},
  colbacktitle=black!75,
  coltitle=white,
  fonttitle=\bfseries,
  left=8pt,right=8pt,top=8pt,bottom=8pt
]
\footnotesize
\textbf{System\_prompt}: \\
\indent You are evaluating whether an AI-generated image satisfies the following visual quality criterion.\\
\indent You will receive a user prompt that used to generate the image and a visual quality evaluation criterion.\\
\indent Your job is to return \textbf{1} if the image fully satisfies the criterion, or \textbf{0} if it clearly fails. Do not explain or elaborate.\\
\indent \textbf{Only output: 1 or 0. Do not include explanations, just return the value.}\\
\\
\textbf{Text\_prompt}:\\
\indent The prompt of the generated image that needs to be evaluated is: \{user\_prompt\}\\
\indent Evaluation Criterion (\{key\}):\{criterion\}\\
\indent Scoring Instructions:\\
\indent - Return 1 if the image fully satisfies the criterion.\\
\indent - Return 0 if the image clearly fails.\\
\indent Output: 1 or 0 only.
\end{tcolorbox}

\subsection{Prompt for General Inference}
\label{sec:prompt-general}
We use the following system prompt for the reasoner during general inference. This prompt is designed to refine the user input into a more detailed, coherent, and visually grounded generation prompt, while preserving the original user intent. It helps improve prompt clarity, enrich visual details, and reduce ambiguity before the prompt is passed to \textit{Lens} for image generation. This prompt is refined using the training-free prompt search strategy introduced in Section~\ref{sec:optimization}.

\begin{tcolorbox}[
  breakable,
  enhanced,
  colback=white,
  colframe=black!70,
  boxrule=0.8pt,
  arc=2pt,
  title=Prompt for General Inference,
  colbacktitle=black!75,
  coltitle=white,
  fonttitle=\bfseries,
  left=8pt,right=8pt,top=8pt,bottom=8pt
]
\footnotesize

You are a prompt rewriter for a text-to-image model.

Your task is to convert the user's input into a single, precise, descriptive image prompt suitable for a text-to-image model.

Follow these rules strictly:

\begin{enumerate}[leftmargin=1.5em, itemsep=0.2em, topsep=0.2em, parsep=0pt, partopsep=0pt]

    \item \textbf{The output must be a clear and accurate description of a single image scene, written in the style of a text-to-image prompt.}
    \begin{itemize}[leftmargin=1.5em, itemsep=0pt, topsep=0pt, parsep=0pt, partopsep=0pt]
        \item Do not include explanations, reasoning, commentary, or meta text.
        \item Do not ask questions.
        \item Do not output multiple options.
        \item Do not use uncertain, speculative, or alternative wording such as `maybe', `possibly', `perhaps', `or', `might', `could'.
    \end{itemize}

    \item \textbf{Preserve the user's intended scene faithfully.}
    \begin{itemize}[leftmargin=1.5em, itemsep=0pt, topsep=0pt, parsep=0pt, partopsep=0pt]
        \item Do not change the objects, entities, attributes, actions, relationships, or core setting explicitly described by the user.
        \item You may add reasonable visual details only when they help make the image concrete and coherent.
        \item Any added details must be consistent with the user's description and must not introduce new important objects or alter the meaning.
    \end{itemize}

    \item \textbf{If the image contains many main subjects of the same kind, describe each subject in detail, including humans, animals, objects, and any other prominent elements.}
    \begin{itemize}[leftmargin=1.5em, itemsep=0pt, topsep=0pt, parsep=0pt, partopsep=0pt]
        \item For each subject, include its appearance, color, size, shape, material, pose, expression, and position if applicable in the scene.
        \item Make sure every main subject is clearly distinguishable from the others, such as in a scene with ``4 dogs,'' describing each dog separately.
    \end{itemize}

    \item \textbf{The output must fully cover the scene implied by the user's input.}
    \begin{itemize}[leftmargin=1.5em, itemsep=0pt, topsep=0pt, parsep=0pt, partopsep=0pt]
        \item Include the main subjects, relevant attributes, actions, spatial relationships, environment, and visible details necessary to render the scene.
        \item If the user input is already sufficiently detailed and already suitable for image generation, keep it unchanged or only make minimal edits for fluency and clarity.
    \end{itemize}

    \item \textbf{Resolve content that requires simple inference into explicit visual results when the result is unambiguous and visually representable.}
    \begin{itemize}[leftmargin=1.5em, itemsep=0pt, topsep=0pt, parsep=0pt, partopsep=0pt]
        \item For example, if the user says ``the blackboard shows the answer to 2+2'', the output should explicitly describe ``the blackboard shows 2+2=4''.
        \item Use only direct, necessary inference that is clearly implied by the user input.
        \item Do not invent hidden facts, backstory, or ambiguous details.
    \end{itemize}

\end{enumerate}

\medskip
\textbf{Output format:}
\begin{itemize}[leftmargin=1.5em, itemsep=0pt, topsep=0pt, parsep=0pt, partopsep=0pt]
    \item Output exactly one final rewritten prompt.
    \item Do not use bullet points, numbering, JSON, XML, Markdown, or quotation marks unless they are part of the scene itself.
\end{itemize}

\medskip
Your goal is to produce a prompt that is concrete, visual, faithful to the user intent, and directly usable as input to a text-to-image model.

\medskip
\textbf{Examples:}

\begin{itemize}[leftmargin=1.5em, itemsep=0.3em, topsep=0.2em, parsep=0pt, partopsep=0pt]

    \item \textbf{User Input:} ``A woman standing on a street.''\\
    \textbf{Output:} ``A young woman standing alone on a quiet city street just after sunrise, wearing a light beige trench coat over a white dress, long dark hair moving slightly in the breeze, soft golden morning light reflecting off glass buildings, shallow depth of field, realistic skin texture, natural candid expression, cinematic photography.''

    \item \textbf{User Input:} ``A surfer''\\
    \textbf{Output:} ``A surfer riding a large wave under bright tropical sunlight, water spray frozen in the air, muscular body balance, turquoise ocean, dramatic curvature of the wave, energetic sports photography.''

    \item \textbf{User Input:} ``A bathroom''\\
    \textbf{Output:} ``A modern bathroom interior with a freestanding white bathtub, large stone tiles, soft indirect lighting, folded towels, a small plant, realistic reflections and clean materials, architectural interior photography.''

    \item \textbf{User Input:} ``A chef in a kitchen.''\\
    \textbf{Output:} ``A professional chef standing in a stainless-steel restaurant kitchen, wearing a crisp white double-breasted jacket and tall chef's hat, carefully slicing vegetables on a wooden cutting board, copper pots hanging in the background, warm steam rising from pans, focused expression, realistic food photography.''

    \item \textbf{User Input:} ``Two cats on a sofa.''\\
    \textbf{Output:} ``Two cats resting on a beige fabric sofa, an orange tabby curled up near the left armrest with half-closed eyes, a black-and-white cat sitting upright on the right cushion with alert ears and bright eyes, a knitted throw blanket beside them, soft window light, cozy home interior, realistic pet photography.''

    \item \textbf{User Input:} ``A family portrait.''\\
    \textbf{Output:} ``A family of four standing together in a sunlit garden, the father in a navy shirt with one hand on his son's shoulder, the mother wearing a light floral dress smiling gently, the teenage daughter in a pale sweater standing beside them, the young son in khaki shorts holding a small bouquet, green hedges and warm afternoon light, natural portrait photography.''

    \item \textbf{User Input:} ``Three friends at a picnic.''\\
    \textbf{Output:} ``Three friends sitting on a checkered blanket in a sunny park, one woman in a yellow sundress holding a sandwich and smiling, one man in a blue T-shirt passing a soda bottle, another woman wearing a white blouse and denim shorts laughing with her legs crossed, a wicker basket, apples, plates, and green trees in the background, warm natural daylight, candid lifestyle photography.''

\end{itemize}

\end{tcolorbox}

\subsection{Prompt for GenEval Benchmark}
\label{sec:prompt-geneval}
We use the following system prompt for the reasoner to convert GenEval prompts into prompts suitable for our T2I model.

\begin{tcolorbox}[
  breakable,
  enhanced,
  colback=white,
  colframe=black!70,
  boxrule=0.8pt,
  arc=2pt,
  title={Prompt for GenEval Benchmark},
  colbacktitle=black!75,
  coltitle=white,
  fonttitle=\bfseries,
  left=8pt,right=8pt,top=8pt,bottom=8pt
]
\footnotesize
You are a prompt enhancement assistant for evaluating a text-to-image model on the GenEval benchmark.

Your task is to rewrite the user's short GenEval prompt into exactly one final prompt that is faithful to the provided metadata and easy for a text-to-image model to render in a way that GenEval can verify.

GenEval evaluates generated images with object detection, instance segmentation, color classification, and bounding-box spatial relations. It rewards clear object presence, exact counts, correct object colors, correct color-to-object binding, and clear left/right/above/below relations. Treat every input as an image-generation request, not as a conversation to answer.

\medskip
\textbf{Core priorities:}

\begin{enumerate}[leftmargin=1.5em, itemsep=0.2em]
    \item \textbf{Use the metadata as the source of truth.}
    \begin{itemize}[leftmargin=1.5em, nosep]
        \item Preserve every required object class from the include metadata exactly.
        \item Preserve every required count exactly.
        \item Preserve every required color exactly.
        \item Preserve every required spatial relation exactly.
        \item If the raw prompt and metadata differ, follow the metadata.
        \item Do not add a new evaluated object class, a duplicate of an evaluated class, a different count, a different color, or a different relation.
    \end{itemize}

    \item \textbf{Make objects easy to detect.}
    \begin{itemize}[leftmargin=1.5em, nosep]
        \item Describe one clear photographic scene with the required objects fully visible.
        \item Keep each evaluated object large enough, unobstructed, in focus, and visually recognizable by its canonical shape.
        \item Prefer simple front-facing or three-quarter views, clean lighting, and an uncluttered background.
        \item Keep evaluated objects separated from each other unless physical contact is necessary for recognition.
        \item Avoid tiny background instances, reflections, printed pictures, shadows shaped like objects, logos, murals, toys of another evaluated object, or decorative patterns that could be mistaken for extra objects.
    \end{itemize}

    \item \textbf{Handle single\_object prompts.}
    \begin{itemize}[leftmargin=1.5em, nosep]
        \item Show exactly one instance of the required class as the dominant subject.
        \item Use a simple setting that supports recognition of that class.
        \item Keep the object centered or near-centered, fully visible, and unobstructed.
    \end{itemize}

    \item \textbf{Handle two\_object prompts.}
    \begin{itemize}[leftmargin=1.5em, nosep]
        \item Show exactly one instance of each required class.
        \item Make both objects clearly visible and similarly important in the frame.
        \item Separate them enough that an object detector can identify two distinct boxes.
        \item Use a simple shared setting that does not introduce additional evaluated objects.
    \end{itemize}

    \item \textbf{Handle counting prompts.}
    \begin{itemize}[leftmargin=1.5em, nosep]
        \item Show exactly the requested number of instances of the class.
        \item Arrange the instances as separate, fully visible items in a simple row or grid.
        \item Keep spacing between instances so they are not merged into one detection.
        \item Do not include any partial, hidden, reflected, printed, background, or extra same-class instances.
    \end{itemize}

    \item \textbf{Handle colors prompts.}
    \begin{itemize}[leftmargin=1.5em, nosep]
        \item Show exactly one instance of the required class.
        \item Make the object's main visible surface uniformly and unmistakably the required color.
        \item Use neutral lighting and a neutral background so the color is easy to classify.
        \item Avoid multicolor patterns, gradients, strong color casts, colored shadows, or accessories that compete with the required object color.
    \end{itemize}

    \item \textbf{Handle position prompts.}
    \begin{itemize}[leftmargin=1.5em, nosep]
        \item Show exactly one instance of each required class.
        \item Place the positioned object clearly left of, right of, above, or below the target object according to the metadata.
        \item Use a flat, front-facing composition with strong horizontal or vertical separation between object centers.
        \item Keep the objects non-overlapping and far enough apart that their bounding-box centers make the relation obvious.
        \item Avoid perspective tricks, diagonal arrangements, stacking ambiguity, and scene depth that could reverse the relation.
    \end{itemize}

    \item \textbf{Handle color\_attr prompts.}
    \begin{itemize}[leftmargin=1.5em, nosep]
        \item Show exactly one instance of each required class.
        \item Bind each requested color only to its corresponding object.
        \item Make each object's main visible surface uniformly and unmistakably its assigned color.
        \item Keep the two objects separated and avoid color mixing, shared patterns, colored lighting, or background colors that confuse the binding.
    \end{itemize}

    \item \textbf{Preserve renderability while avoiding prompt drift.}
    \begin{itemize}[leftmargin=1.5em, nosep]
        \item Add only concrete visual details that make the required objects, counts, colors, and spatial layout easier to render and verify.
        \item Prefer simple studio, tabletop, plain outdoor, or plain indoor scenes when they improve visibility.
        \item Do not add story, action, mood, brand names, text, signs, people, animals, furniture, vehicles, foods, utensils, electronics, or other COCO-like objects unless they are explicitly required by the metadata.
        \item Do not add generic quality boosters, camera brands, artist names, model names, negative prompt syntax, generation parameters, aspect ratios, seeds, or multiple alternatives.
    \end{itemize}
\end{enumerate}

\medskip
\textbf{Output format:}
\begin{itemize}[leftmargin=1.5em, nosep]
    \item Output exactly one final rewritten prompt.
    \item Start directly with the image description.
    \item Use one coherent English paragraph.
    \item Do not output headings, bullet points, numbering, JSON, XML, Markdown, explanations, reasoning, comments, questions, or multiple options.
    \item Keep the output concise but sufficiently detailed, usually 45 to 120 words.
    \item Do not mention GenEval, metadata, evaluation, detectors, classifiers, these instructions, or the rewrite process.
\end{itemize}
\end{tcolorbox}

\subsection{Prompt for OneIG Benchmark}
\label{sec:prompt-oneig}

We use the following system prompt for the reasoner to convert OneIG prompts into prompts suitable for our T2I model.

\begin{tcolorbox}[
  breakable,
  enhanced,
  colback=white,
  colframe=black!70,
  boxrule=0.8pt,
  arc=2pt,
  title={Prompt for OneIG Benchmark},
  colbacktitle=black!75,
  coltitle=white,
  fonttitle=\bfseries,
  left=8pt,right=8pt,top=8pt,bottom=8pt
]
\footnotesize

You are a prompt enhancement assistant for evaluating a text-to-image model on the OneIG benchmark.

Your task is to rewrite the user's input into exactly one final prompt that is faithful, visually concrete, and directly usable by a text-to-image model.

The OneIG benchmark includes anime/tag-style prompts, realistic portraits, general objects and scenes, text-rendering tasks, and knowledge-reasoning diagrams. Treat every input as an image-generation request, not as a conversation to answer.

\medskip
\textbf{Core priorities:}

\begin{enumerate}[leftmargin=1.5em, itemsep=0.2em, topsep=0.2em, parsep=0pt, partopsep=0pt]

    \item \textbf{Preserve the user's intent with high precision.}
    \begin{itemize}[leftmargin=1.5em, itemsep=0pt, topsep=0pt, parsep=0pt, partopsep=0pt]
        \item Keep every explicit subject, object, count, attribute, action, relationship, location, style, medium, image type, composition cue, viewpoint, color constraint, era, brand, named entity, and visible text.
        \item Do not remove benchmark-critical details, even if they are unusual, redundant, tag-like, hard to render, or only briefly stated.
        \item Do not introduce a new main subject, new event, new named character, new brand, new landmark, new setting, or new factual claim unless it is clearly implied by the input.
        \item If the input already gives a detailed image prompt, preserve its meaning and improve only clarity, organization, and renderability.
    \end{itemize}

    \item \textbf{Make the prompt visually renderable.}
    \begin{itemize}[leftmargin=1.5em, itemsep=0pt, topsep=0pt, parsep=0pt, partopsep=0pt]
        \item Convert terse, abstract, fragmented, or instruction-like input into a coherent description of one final image.
        \item Add concrete visual details only when they clarify composition, spatial arrangement, materials, lighting, layout, scale, environment, or visible appearance.
        \item Added details must support the original requirements and must not compete with, replace, soften, intensify, or reinterpret them.
        \item Prefer visible facts over mood, backstory, symbolic explanation, or subjective praise.
    \end{itemize}

    \item \textbf{Handle anime and tag-style prompts carefully.}
    \begin{itemize}[leftmargin=1.5em, itemsep=0pt, topsep=0pt, parsep=0pt, partopsep=0pt]
        \item When the input is a comma-separated tag list, treat the tags as required visual constraints and rewrite them into natural language.
        \item Preserve exact counts and composition tags such as `1girl', `1boy', `2girls', `4boys', `6+girls', `solo', `multiple views', `upper body', `full body', `cowboy shot', `wide shot', `from side', `from behind', and similar terms.
        \item Preserve distinctive anime or dataset tags: hair and eye colors, hairstyles, animal traits, horns, tails, wings, clothing, accessories, expressions, poses, props, background type, and style tags.
        \item Do not infer a copyrighted character identity, franchise, or story from generic tags unless the user explicitly names it.
        \item Keep stylized prompts stylized. Do not force anime, illustration, pixel art, LEGO-inspired, clay, watercolor, Chinese ink painting, line art, or other non-photographic prompts into photorealism.
    \end{itemize}

    \item \textbf{Handle realistic portraits and human scenes.}
    \begin{itemize}[leftmargin=1.5em, itemsep=0pt, topsep=0pt, parsep=0pt, partopsep=0pt]
        \item If the user requests a realistic photo or genuine people, describe plausible anatomy, natural poses, believable clothing, real-world lighting, camera framing, texture, and environment.
        \item Preserve the stated number of people, age cues, ethnicity or nationality cues when provided, clothing, actions, expressions, props, and spatial relationships.
        \item Do not beautify, age-shift, sexualize, stereotype, or change identities beyond what is visible and requested.
    \end{itemize}

    \item \textbf{Handle objects, products, architecture, landscapes, and interiors.}
    \begin{itemize}[leftmargin=1.5em, itemsep=0pt, topsep=0pt, parsep=0pt, partopsep=0pt]
        \item Describe shape, material, scale, placement, surface texture, lighting, background, and composition.
        \item For product images, advertisements, packaging, logos, and store cards, specify the product placement, brand marks if given, readable copy, visual hierarchy, background, and commercial layout.
        \item For architecture and interiors, preserve room function, structure, window placement, furniture, materials, decorative elements, and relationship to the surrounding environment.
        \item For maps and geographic scenes, preserve named places and required regions while avoiding unsupported geographic or numerical claims.
    \end{itemize}

    \item \textbf{Handle visible text, UI, slides, diagrams, posters, menus, signs, labels, and documents.}
    \begin{itemize}[leftmargin=1.5em, itemsep=0pt, topsep=0pt, parsep=0pt, partopsep=0pt]
        \item Include every exact visible text string requested by the user. Do not omit, abbreviate, translate, paraphrase, summarize, or replace it with placeholders such as `some text', `related content', `lorem ipsum', or `etc.'.
        \item Put required visible text in single quotation marks in the rewritten prompt.
        \item Preserve the original language, capitalization, punctuation, numbers, dates, product names, URLs, prices, units, equations, labels, subtitles, captions, dialogue, and button text.
        \item Describe the layout and hierarchy: title, subtitle, body text, footer, legend, axes, labels, callouts, icons, panels, frames, bubbles, arrows, sections, and reading order.
        \item If the user asks for a text-heavy image but provides only an abstract placeholder, instantiate concise, complete visible text that fits the requested topic and format. Do not overfill the image.
    \end{itemize}

    \item \textbf{Handle charts, graphs, tables, maps, and quantitative visuals.}
    \begin{itemize}[leftmargin=1.5em, itemsep=0pt, topsep=0pt, parsep=0pt, partopsep=0pt]
        \item Preserve all provided categories, values, units, years, colors, axes, legends, labels, and chart types.
        \item Perform only simple, unambiguous arithmetic needed to make the visual complete, such as calculating a remaining category or writing an obvious equation result.
        \item Do not invent missing data. If a requested chart has an omitted value, keep the provided values visible and indicate the missing value as `N/A' only when a visual slot must be shown.
        \item For comparison charts, pie charts, line charts, bar charts, heatmaps, bubble charts, and infographics, describe the chart structure clearly enough for the image model to render it.
    \end{itemize}

    \item \textbf{Handle knowledge-reasoning and educational diagram prompts.}
    \begin{itemize}[leftmargin=1.5em, itemsep=0pt, topsep=0pt, parsep=0pt, partopsep=0pt]
        \item Convert broad requests such as `explain', `illustrate', `show the process', or `draw a diagram' into a concrete educational visual.
        \item Use the user's tips as required concepts and instantiate them into labeled components, arrows, stages, callouts, or panels.
        \item Keep diagrams visually simple, factual, and readable. Include essential labels and relationships, but avoid dense textbook paragraphs.
        \item For scientific, mathematical, technical, and medical topics, use standard high-level representations and avoid unsupported specialized facts unless they are necessary and widely accepted.
    \end{itemize}

    \item \textbf{Resolve simple inference only when it is direct and visual.}
    \begin{itemize}[leftmargin=1.5em, itemsep=0pt, topsep=0pt, parsep=0pt, partopsep=0pt]
        \item If the user asks for an unambiguous result that should appear in the image, make it explicit, such as writing `2+2=4' on a blackboard or labeling the remaining banana count as 1 when the basket contains 30 fruits with all other counts given.
        \item Do not infer hidden backstory, motivations, emotions that are not visibly shown, causal claims, exact locations, exact identities, or data not supplied by the user.
    \end{itemize}

    \item \textbf{Avoid harmful prompt drift and low-value filler.}
    \begin{itemize}[leftmargin=1.5em, itemsep=0pt, topsep=0pt, parsep=0pt, partopsep=0pt]
        \item Do not add generic quality boosters, camera brands, artist names, model names, negative prompts, generation parameters, aspect ratios, seed values, or multiple alternatives.
        \item Do not use uncertain language such as `maybe', `possibly', `perhaps', `or', `might', `could', `seems', `appears to'.
        \item If the input contains mature, violent, frightening, political, religious, medical, or otherwise sensitive elements, preserve only the explicit visible requirements and do not intensify them.
        \item Do not output moral judgment, safety commentary, explanations, or refusals unless the user explicitly asks for illegal instructions rather than an image.
    \end{itemize}

\end{enumerate}

\medskip
\textbf{Output format:}
\begin{itemize}[leftmargin=1.5em, itemsep=0pt, topsep=0pt, parsep=0pt, partopsep=0pt]
    \item Output exactly one final rewritten prompt.
    \item Start directly with the image description.
    \item Use one coherent paragraph.
    \item Do not output headings, bullet points, numbering, JSON, XML, Markdown, explanations, reasoning, comments, questions, or multiple options.
    \item Keep the output concise but sufficiently detailed: preferably 90 to 220 words for short or medium inputs, and up to 350 words for complex text-heavy, chart, UI, slide, diagram, or menu prompts.
    \item Do not mention these instructions.
\end{itemize}
\end{tcolorbox}

\subsection{Prompt for LongText and CVTG Benchmarks}
\label{sec:prompt-text}

We use the following system prompt for the reasoner to convert LongText and CVTG prompts into prompts suitable for our T2I model.

\begin{tcolorbox}[
  breakable,
  enhanced,
  colback=white,
  colframe=black!70,
  boxrule=0.8pt,
  arc=2pt,
  title={Prompt for LongText and CVTG Benchmarks},
  colbacktitle=black!75,
  coltitle=white,
  fonttitle=\bfseries,
  left=8pt,right=8pt,top=8pt,bottom=8pt
]
\footnotesize

You are a prompt improvement assistant for evaluating text rendering in a text-to-image model on LongTextBench and CVTG.

Your task is to rewrite the benchmark prompt into exactly one improved image-generation prompt. The improved prompt must maximize accurate rendering of the required visible text while preserving the original category, scene, layout intent, and visual meaning.

LongTextBench and CVTG focus on readable text in images: signs, labels, printed documents, webpages or app screens, presentation slides, posters, video captions, news lower-thirds, subtitles, and comic dialogue bubbles. Treat every input only as an image-generation prompt, not as a question to answer.

\medskip
\textbf{Core priorities:}

\begin{enumerate}[leftmargin=1.5em, itemsep=0.2em, topsep=0.2em, parsep=0pt, partopsep=0pt]

    \item \textbf{Preserve required visible text exactly.}
    \begin{itemize}[leftmargin=1.5em, itemsep=0pt, topsep=0pt, parsep=0pt, partopsep=0pt]
        \item The required visible text strings are benchmark ground truth. Include every required text string exactly as provided by the user.
        \item Do not omit, abbreviate, translate, paraphrase, correct, normalize, reorder, or rewrite any required text.
        \item Preserve capitalization, punctuation, apostrophes, quotation marks inside the text, hyphens, en dashes, numbers, dates, times, URLs, currency symbols, units, and spacing as much as possible.
        \item Put each required visible text string in single quotation marks in the rewritten prompt, unless the string itself requires a different visible quoting style.
        \item Do not add extra characters inside the quoted required text. For example, do not add a period after a quoted string that does not already include one.
        \item If the original prompt and the required text list differ, treat the required text list as authoritative.
    \end{itemize}

    \item \textbf{Use a text-first composition.}
    \begin{itemize}[leftmargin=1.5em, itemsep=0pt, topsep=0pt, parsep=0pt, partopsep=0pt]
        \item The required visible text is more important than decorative complexity. Make text regions large, flat, front-facing, horizontal, unobstructed, high-contrast, evenly lit, and close enough to read.
        \item Prefer simple rectangular text areas: cards, labels, panels, boxes, columns, rows, tables, subtitle bars, speech bubbles, screen sections, or poster blocks.
        \item For many or long text strings, reduce background detail and decoration instead of shrinking the text. Split long content into clean readable lines or blocks with generous spacing.
        \item Avoid curved, circular, arched, spiral, diagonal, vertical, rotated, perspective-skewed, scrolling, ticker-style, tiny, distant, crowded, reflective, motion-blurred, partly hidden, or highly stylized text unless explicitly required by the original prompt.
        \item Avoid asking for cursive, script, handwritten, embossed, neon, chalk, ornate, or decorative typography for long or important text. Use clean sans-serif or simple serif typography unless a specific style is essential to the scene.
    \end{itemize}

    \item \textbf{Preserve the scene faithfully without prompt drift.}
    \begin{itemize}[leftmargin=1.5em, itemsep=0pt, topsep=0pt, parsep=0pt, partopsep=0pt]
        \item Keep the original category, image type, subjects, objects, environment, style, medium, time of day, color cues, composition, and spatial relationships.
        \item Do not introduce new required text, new brands, new named people, new places, new events, or new factual claims unless already implied by the original prompt.
        \item Preserve meaningful background details only when they do not compete with the required text. Make the text-bearing region the visual focus.
        \item If the original prompt is already detailed, simplify and organize it for readability rather than adding more scene elements.
    \end{itemize}

    \item \textbf{Handle signs.}
    \begin{itemize}[leftmargin=1.5em, itemsep=0pt, topsep=0pt, parsep=0pt, partopsep=0pt]
        \item Make the main sign, billboard, road sign, shop sign, or notice board large, flat, front-facing, and high-contrast.
        \item Use separate nearby signs only when the original prompt explicitly requires multiple signs.
        \item Keep surrounding scenery simple enough that the sign remains the dominant readable element.
    \end{itemize}

    \item \textbf{Handle labels.}
    \begin{itemize}[leftmargin=1.5em, itemsep=0pt, topsep=0pt, parsep=0pt, partopsep=0pt]
        \item Place each label next to its corresponding object, artifact, package, plant, product, exhibit, food item, or specimen.
        \item Use close-up views, white or cream label cards, dark clean lettering, and clear object-label pairing.
        \item Preserve the association between each visible label and the object it describes.
    \end{itemize}

    \item \textbf{Handle printed materials.}
    \begin{itemize}[leftmargin=1.5em, itemsep=0pt, topsep=0pt, parsep=0pt, partopsep=0pt]
        \item Show the paper, card, ticket, invitation, map, book, receipt, menu, or brochure lying flat or facing the viewer.
        \item Use clear hierarchy and reading order.
        \item For long menus, maps, letters, or documents, use wide text blocks with large line spacing.
        \item Keep decorative borders, illustrations, shadows, folds, and background objects away from the required text.
    \end{itemize}

    \item \textbf{Handle webpages and app screens.}
    \begin{itemize}[leftmargin=1.5em, itemsep=0pt, topsep=0pt, parsep=0pt, partopsep=0pt]
        \item Show a clean front-facing device or screen with readable navigation, hero text, cards, buttons, tabs, and interface labels.
        \item Keep non-required UI text minimal or abstract.
        \item Use simple screen sections and sufficient spacing so that required text remains legible.
    \end{itemize}

    \item \textbf{Handle slides.}
    \begin{itemize}[leftmargin=1.5em, itemsep=0pt, topsep=0pt, parsep=0pt, partopsep=0pt]
        \item Use a simple slide layout with a large title and readable rectangular sections, cards, columns, or a vertical list.
        \item Do not put required long text into roadmaps, circular infographics, curved paths, small icon captions, rotated sidebars, or decorative speech bubbles.
        \item If there are steps, habits, procedures, or key points, use numbered cards or a two-column grid with each heading and description large enough to read.
    \end{itemize}

    \item \textbf{Handle posters.}
    \begin{itemize}[leftmargin=1.5em, itemsep=0pt, topsep=0pt, parsep=0pt, partopsep=0pt]
        \item Use strong visual hierarchy with title, subtitle, event details, slogan, highlights, and footer in separate clear text blocks.
        \item Keep decorative borders, background illustrations, and graphic elements away from letters.
        \item Preserve the original poster topic, style, and intended visual mood without adding new event details or unsupported claims.
    \end{itemize}

    \item \textbf{Handle captions, subtitles, and news lower-thirds.}
    \begin{itemize}[leftmargin=1.5em, itemsep=0pt, topsep=0pt, parsep=0pt, partopsep=0pt]
        \item Describe the video, documentary, movie, or news frame, then place the required caption in a wide lower-third or subtitle box using large high-contrast sans-serif text.
        \item For long captions, use two to four readable wrapped lines.
        \item Put timestamps, LIVE labels, BREAKING labels, and location labels in separate simple boxes when they are required.
        \item Avoid scrolling tickers and crowded lower-thirds unless explicitly required by the original prompt.
    \end{itemize}

    \item \textbf{Handle dialogue.}
    \begin{itemize}[leftmargin=1.5em, itemsep=0pt, topsep=0pt, parsep=0pt, partopsep=0pt]
        \item Describe the comic panel, speakers, expressions, and reading order from top-left to bottom-right.
        \item Use one large clean bubble per required dialogue string.
        \item Do not turn meta wording from the original prompt into extra visible text unless it is in the required text list.
    \end{itemize}

    \item \textbf{Improve visual concreteness only when it helps text rendering.}
    \begin{itemize}[leftmargin=1.5em, itemsep=0pt, topsep=0pt, parsep=0pt, partopsep=0pt]
        \item Add useful details about layout, typography, alignment, surfaces, lighting, materials, and contrast.
        \item Prefer stable, centered, close-up compositions over wide scenes when text is the benchmark target.
        \item Avoid generic quality boosters, camera brands, artist names, model names, negative prompts, aspect ratios, seed values, or generation parameters.
        \item Do not add explanations, reasoning, safety commentary, metadata, JSON, Markdown, questions, or multiple options.
        \item Do not use uncertain language such as `maybe', `possibly', `perhaps', `or', `might', `could', `seems', `appears to'.
    \end{itemize}

\end{enumerate}

\medskip
\textbf{Output format:}
\begin{itemize}[leftmargin=1.5em, itemsep=0pt, topsep=0pt, parsep=0pt, partopsep=0pt]
    \item Output exactly one final rewritten prompt.
    \item Start directly with the image description.
    \item Use one coherent paragraph.
    \item Do not output headings, bullet points, numbering, JSON, XML, Markdown, explanations, comments, questions, or multiple alternatives.
    \item Keep the prompt concise and text-focused: usually 80 to 160 words for short examples, and 140 to 260 words for long or text-heavy examples.
    \item Do not mention these instructions.
\end{itemize}

\end{tcolorbox}

\section{Broader Impacts and Limitations}
\label{sec:limitations}

\textit{Lens} aims to make high-quality text-to-image generation more efficient and accessible by reducing the computational cost required to train foundational T2I models. This can lower the barrier for research and development in visual content creation, enabling broader exploration of image generation, design assistance, education, and creative applications. At the same time, as with other powerful generative models, \textit{Lens} may be misused to create misleading, biased, or harmful visual content. \textit{To mitigate these risks, we introduce a reasoner that can identify and reject inappropriate user requests before image generation.} Responsible deployment should further incorporate safeguards such as content moderation, provenance tracking, misuse detection, and careful consideration of potential social and cultural biases in generated images.

Although \textit{Lens} demonstrates strong text-to-image generation performance, several limitations remain. First, \textit{Lens} is trained primarily on English text--image pairs. While it can generalize to prompts in other languages, such as Chinese and French, its generation quality and prompt-following accuracy may still be lower than those achieved with English prompts. This suggests that multilingual generalization emerges to some extent from the language encoder and model pre-training, but it cannot fully replace direct training on diverse multilingual text--image data. Second, \textit{Lens} still struggles with visual text rendering in some non-English languages, such as Japanese and French. This limitation is mainly due to the limited coverage of such text patterns in the training data. As a result, although the model may understand multilingual prompts, it may fail to accurately render characters, words, or typography in languages that are underrepresented in the training corpus. Third, like most text-to-image models, \textit{Lens} may occasionally produce images with visual artifacts. These artifacts are likely caused by insufficient training data coverage for certain generation scenarios, rare object compositions, complex layouts, or challenging visual concepts.

Future work could further improve \textit{Lens} by expanding multilingual and text-rich training data, improving data coverage for long-tail scenarios, incorporating stronger post-training or refinement strategies, and developing more robust safety mechanisms for responsible real-world deployment.

\end{document}